\documentclass[10pt,twocolumn,letterpaper]{article}

\usepackage[pagenumbers]{cvpr} 

\usepackage{graphicx}
\usepackage{amsmath}
\usepackage{amssymb}
\usepackage{booktabs}
\usepackage{diagbox}
\usepackage{comment}

\usepackage[pagebackref,breaklinks,colorlinks]{hyperref}

\usepackage[capitalize]{cleveref}
\crefname{section}{Sec.}{Secs.}
\Crefname{section}{Section}{Sections}
\Crefname{table}{Table}{Tables}
\crefname{table}{Tab.}{Tabs.}

\def\cvprPaperID{11759} 
\def\confName{CVPR}
\def\confYear{2022}

\begin{document}

\title{Towards Precision in Appearance-based Gaze Estimation in the Wild}
\author{Murthy L.R.D.\\
Indian Institute of Science, Bangalore\\
{\tt\small lrdmurthy@iisc.ac.in}
\and
Abhishek Mukhopadhyay\\
Indian Institute of Science, Bangalore\\
{\tt\small abhishekmukh@iisc.ac.in}
\and
Shambhavi Aggarwal\\
GCET, Vallabh Vidyanagar, India\\
{\tt\small agg.shambhavi@gmail.com}
\and
Ketan Anand\\
 M. S. Ramaiah Institute of Technology, India\\
{\tt\small anand\_ketan@outlook.com}
\and
Pradipta Biswas\\
Indian Institute of Science, Bangalore\\
{\tt\small pradipta@iisc.ac.in}
}
\maketitle
\begin{abstract}
Appearance-based gaze estimation systems have shown great progress recently, yet the performance of these techniques depend on the datasets used for training. Most of the existing gaze estimation datasets setup in interactive settings were recorded in laboratory conditions and those recorded in the wild conditions display limited head pose and illumination variations. Further, we observed little attention so far towards precision evaluations of existing gaze estimation approaches.  In this work, we present a large gaze estimation dataset, PARKS-Gaze, with wider head pose and illumination variation and with multiple samples for a single Point of Gaze (PoG). The dataset contains 974 minutes of data from 28 participants with a head pose range of $\pm60^{\circ}$ in both yaw and pitch directions. Our within-dataset and cross-dataset evaluations and precision evaluations indicate that the proposed dataset is more challenging and enable models to generalize on unseen participants better than the existing in-the-wild datasets. The project page can be accessed here: https://github.com/lrdmurthy/PARKS-Gaze

\end{abstract}

\section{Introduction}
\label{sec:intro}
Imagine how cool would it be if the mouse cursor on your laptop screen moves wherever you see. Appearance-based gaze estimation systems make this possible just with built-in webcams by obviating the need of infra-red (IR) illuminators and multiple camera systems. But how do you feel about the same technology when the cursor makes sudden, unexpected movements and jumps around that specific icon you are staring to select. Precision indicates the consistency of a model’s gaze predictions and is an imperative feature to build gaze-controlled applications. Precise gaze estimates allow not only for smooth interaction experience, but also useful to improve the accuracy with the help of person-specific calibration. This paper investigates this long-ignored aspect, “precision” of appearance-based gaze estimation systems and attempts to improve it.


Existing research shows that numerous applications were built using appearance-based gaze estimation systems with built-in mobile phone cameras \cite{zhang2017smartphone}, laptop cameras \cite{murthy2021i2dnet} and with commodity web cameras for various contexts like automotive \cite{lrd2022distraction, lrd2022efficient}, assistive technology \cite{sharma2020webcam} and human-robot interaction \cite{krishna2022comparing}. Recent studies \cite{zhang2019evaluation, murthy2022deep} have shown the limitations of highly accurate IR-based gaze estimation systems and highlighted the robustness of appearance-based gaze estimation systems under similar operating conditions. Since appearance-based gaze estimation systems are predominantly learning-based approaches, the performance of such models under various scenarios is influenced by the diversity and characteristics of the training dataset. 

Several appearance-based gaze estimation datasets like EYEDIAP \cite{funes2014eyediap}, ColumbiaGaze \cite{smith2013gaze}, UT-Multiview \cite{sugano2014learning} and ETH-X Gaze \cite{zhang2020eth} were recorded in constrained laboratory settings. However, a model trained with a dataset collected in such restricted scenarios and due to lack of diverse scenarios like \textit{ambient illumination, head poses, inter-person variations, intra-person appearance variations like facial hair, makeup and facial expressions}, will be less accurate when deployed in the real world  
We observed only three gaze estimation datasets \cite{krafka2016eye, zhang2017mpiigaze, kellnhofer2019gaze360} were collected outside laboratory settings so far. Even though GazeCapture involved large number of participants, it had limited gaze angle distribution and Zhang et al. \cite{zhang2017s} reported that MPIIFaceGaze contained limited head pose. Authors of Gaze360 \cite{kellnhofer2019gaze360} commented that GazeCapture contains outdoor scenarios partially and MPIIFaceGaze was recorded indoor. Moreover, MPIIFaceGaze \cite{zhang2017s} was recorded in static settings where artifacts like motion blur were not considered which arises in dynamic head movement settings. 

For applications like gaze-controlled interfaces, dwell-time based selection techniques are widely used and allows for a hands-free interaction experience to the users. Such dwell-time based systems relies on a statically defined cumulative fixation (dwell) duration to trigger the click/selection event \cite{nayyar2017optidwell}. In order to utilize such dwell-time based techniques, the gaze estimation system must provide precise gaze estimates. But, we did not observe such precision analysis of any existing gaze estimation model in the literature. 

This paper presents PARKS-Gaze, a large appearance-based gaze estimation dataset collected in both indoor and outdoor scenarios with dynamic head movement and wider head pose than existing in-the-wild datasets. Further, our proposed dataset contained large intra-person appearance variations. It also contained large number of image samples per gaze target with appearance and head pose variations to facilitate the precision analysis. We obtained data samples across wide illumination variations with a head pose range of $\pm60^{\circ}$ in both yaw and pitch directions. We evaluated the proposed dataset using within and cross-dataset evaluations and observed that it is more challenging and enable the models to generalize better on unseen participants than existing in-the-wild datasets. The proposed dataset extended the dataset presented by Murthy et al. \cite{lrd2022parks} and the significant additions in the present work are listed below.
\begin{itemize}
\item The current version of dataset contains 10 more participants and 65\% more data samples than earlier version.
\item A within-dataset evaluation is performed using existing AGE-Net, I2D-Net and Dilated-Net models. 
\item A cross-dataset evaluation including GazeCapture \cite{krafka2016eye} and ETH-X Gaze \cite{zhang2020eth} datasets in addition to other datasets is presented and observed superior results.
\item Effect of gaze estimation models and training datasets on gaze precision is presented. 
\end{itemize}

\begin{table*}
\centering
  \caption{A Brief summary of Gaze estimation Datasets}
  \label{datasets_summary}
  \begin{tabular}{c|c|c|c|c|c|c|c} 
    \hline
    \toprule
    Dataset & Participants & Head Pose  & Data & Region & Temporal Data & Setting \\
    \midrule
    \midrule
    ColumbiaGaze \cite{smith2013gaze} & 56 & 0, $\pm$30$^{\circ}$  & 5,880 & Frame & - & Lab \\
    UT-Multiview \cite{sugano2014learning}  & 50 & $\pm$36$^{\circ}$, $\pm$36$^{\circ}$ & 64,000 & Eyes & - & Lab\\
    EYEDIAP \cite{funes2014eyediap}  & 16  & $\pm$25$^{\circ}$, 30$^{\circ}$  & 237 Min. & Frame & 30 Hz & Lab\\
    RT-GENE \cite{fischer2018rt} & 15 & $\pm$40$^{\circ}$, -40$^{\circ}$ & 122,531 & Face + Eyes &  - & Lab \\
    ETH-XGaze \cite{zhang2020eth} & 110 & $\pm$80$^{\circ}$, $\pm$80$^{\circ}$ & $\approx 1.1M $ & Face & - & Lab \\
    \hline
    GazeCapture \cite{krafka2016eye}  & 1474 & $\pm$30$^{\circ}$, [-20$^{\circ}$, 40$^{\circ}$] & $\approx2.1M$ &  Frame &  - & Daily Life  \\
    MPIIGaze \cite{zhang2017mpiigaze} & 15 & $\pm$25$^{\circ}$, [-10$^{\circ}$,30$^{\circ}$] & 213,659 &  Eyes &  - & Daily Life\\
    MPIIFaceGaze \cite{zhang2017s} & 15 & $\pm$25$^{\circ}$, [-10$^{\circ}$,30$^{\circ}$]  & 37,667 & Face & - & Daily Life\\    \hline
    \hline
\textbf{PARKS-Gaze} & \textbf{28} & \textbf{$\pm$60$^{\circ}$, $\pm$60$^{\circ}$} & \textbf{300,961} & \textbf{Face + Eyes} & 6 Hz & \textbf{Daily Life} \\ 
  \bottomrule
\end{tabular}
\end{table*}

\section{Related Work}\label{LitReview}
\subsection{Datasets}
Existing appearance-based gaze estimation datasets can be broadly divided into two categories. (I) Datasets recorded in controlled environment; (II) Datasets recorded in real-world scenarios. We briefly summarized the existing datasets in Table \ref{datasets_summary}. 
In the 'Data' column of Table \ref{datasets_summary}, the numbers represent the number of data samples available unless specified otherwise. 

\subsubsection{\textbf{Datasets - Controlled Conditions}:}
ColumbiaGaze \cite{smith2013gaze} was one of the first datasets created for eye gaze estimation recorded in laboratory settings. Even though the number of images (6K) were less than most of the existing datasets, it included large number of participants (N=58). 
Following ColumbiaGaze, UTMultiview \cite{sugano2014learning} was proposed with 50 participants and 1.1M images which contained both natural eye images captured from multiple cameras and synthetic eye images. The head pose was fixed while recording UTMultiview and ColumbiaGaze captured 5 head poses from their respective participants. Subsequent datasets focused on capturing gaze data with unconstrained head pose. EYEDIAP \cite{funes2014eyediap} proposed one of the first video datasets for gaze estimation. 
EYEDIAP contained 94 videos, each of length 2-3 minutes, recorded with 16 participants. Huang et al. \cite{huang2017tabletgaze} proposed TabletGaze, a 2D gaze estimation dataset with 51 participants using Tablet PCs with 4 holding postures and 35 gaze locations. 
These datasets achieved limited gaze area by virtue of the screens and Tablet PCs used to record the data. We observed that subsequent datasets attempted to overcome this challenge in multiple ways. RT-GENE \cite{fischer2018rt} utilized mobile eye tracking glasses and a free-viewing approach for the participants who were positioned at a distance of 0.5 to 2.9 meters from the camera to achieve wider gaze region. 
Since RT-GENE dataset's face images were captured using mobile eye tracking glasses, they proposed semantic inpainting approach to remove the eye tracking glasses but \cite{cheng2020gaze} reported noise in the in-painted images. The most recent ETH-XGaze \cite{zhang2020eth} dataset presented over 1 million images recorded with diverse head pose and gaze angle distribution. ETH-XGaze was recorded in laboratory conditions with 110 participants and they focused on capturing high resolution images with gaze angles under extreme head poses. 
All the above discussed datasets were recorded in controlled conditions in lab where illumination 
and appearance of the participant had little variation. These datasets did not capture the appearance of participants in the wild conditions and due to predefined recording setups, they failed to capture intra-person appearance variations.
\subsubsection{\textbf{Datasets - Real World Scenarios}:}
We identified three gaze estimation datasets which were recorded in daily life scenarios. Krafka et al. proposed GazeCapture \cite{krafka2016eye} dataset 
with 1,474 participants and 2.4M images were captured using mobile devices and authors provided corresponding 2D gaze annotations. Zhang et al. proposed MPIIGaze \cite{zhang2017mpiigaze} with 3D gaze annotations, where a total of 213,659 eye images were provided from 15 participants using laptop devices in daily life conditions. Further, authors released MPIIFaceGaze \cite{zhang2017s}, a subset of MPIIGaze with 37667 original face images. The evaluation set of MPIIFaceGaze contained a total of 45K images, obtained by flipping a portion of images horizontally. This dataset covered diverse scenarios of illumination and appearance. 
Since the gaze direction range was constrained by laptops or mobile devices in both MPIIGaze and GazeCapture, Kellnhofer et al. proposed Gaze360 \cite{kellnhofer2019gaze360}, a large scale dataset recorded using AprilTag markers and a 360$^{\circ}$ camera to acquire unconstrained gaze data. Gaze360 was recorded with 238 participants in 5 indoor and 2 outdoor locations and captured a total of 172K images. Further, the head pose distribution of Gaze360 is unknown \cite{zhang2020eth} while having limited samples per-person and hence limited intra-person appearance variations. 
Since we aim to build a dataset for gaze-controlled interfaces, our work is closely related to MPIIFaceGaze and GazeCapture datasets which were recorded in daily life scenarios and we primarily consider these for further comparisons. 

\subsection{Gaze Estimation Models}

We can broadly classify the existing gaze estimation models into Single-Channel and Multi-channel architectures. One of the first attempts was GazeNet \cite{zhang2017mpiigaze}, a single channel approach where a single eye image was used as the input to an architecture based on VGG-16. 
GazeNet reported a 5.4$^{\circ}$ mean angle error on MPIIGaze. This work was followed by Spatial-Weights CNN \cite{zhang2017s} where full face images were provided as input. Later, multi-channel approaches were proposed as an alternative to single-channel approaches. iTracker \cite{krafka2016eye} was one of the first multi-channel architecture which used left eye image, right eye image, face crop image and face grid information as inputs. Multi-Region Dilated-Net proposed by Chen and Shi \cite{chen2018appearance} employed dilated convolutions in their proposed model and used both eye images along with face image as inputs. This approach also reported 4.8$^{\circ}$ mean angle error as \cite{zhang2017s} did on MPIIFaceGaze. Cheng et al. proposed FAR-Net \cite{cheng2020gaze} which utilized the asymmetry between two eyes of same person to obtain gaze estimates. In this work, they generated confidence scores for the gaze estimates obtained from two eye images to choose the more accurate prediction. Murthy et al. \cite{murthy2021i2dnet} proposed an approach for gaze estimation based on difference mechanism and reported 4.3$^{\circ}$ and 8.44$^{\circ}$ accuracy on MPIIFaceGaze and RT-GENE datasets respectively. The difference mechanism utilized the absolute difference of the feature vectors computed from both eye images for gaze estimation. Cheng et al. \cite{cheng2020coarse} proposed a coarse to fine approach for building a gaze estimation method and reported 4.14$^{\circ}$ and 5.3$^{\circ}$ accuracy on MPIIFaceGaze and EYEDIAP datasets respectively. Recently, Murthy and Biswas proposed AGE-Net \cite{murthy2021appearance} which utilized feature-level attention mechanism to obtain a 4.09$^{\circ}$ and 7.44$^{\circ}$ error on MPIIFaceGaze and RT-GENE datasets respectively. 

In summary, we observed that only two appearance-based gaze estimation datasets were collected in real world scenarios under interactive setting. Among them, MPIIFaceGaze \cite{zhang2017s} was recorded under static settings with limited head pose distribution and GazeCapture \cite{krafka2016eye} contained limited gaze direction distribution. Further, no gaze estimation model was evaluated for its precision. 

\section{PARKS-Gaze - Dataset}\label{OurDat}

The main objective of creating this dataset is to capture all naturally possible head pose appearances from participants while interacting with the display. We also wanted to capture user's natural facial expressions and appearance variations during gaze data recordings, emulating the natural interaction scenarios. We captured large number of images for a single PoG which enable researchers to compute gaze estimation models' precision along with accuracy. For this purpose, participants' laptop devices were used and in-built cameras were utilized as the recording devices since it facilitated us to record in any location in unconstrained manner. Since the focus is on interactive applications like gaze-controlled interfaces, laptop devices were chosen in spite of the resultant limited gaze direction range. We recruited 28 participants and installed our custom data recording software on their laptops. We explained each participant about our experiment and obtained all necessary permissions and consent.

\subsection{Data Collection Procedure}
Once a recording session was initiated by the participant, our custom software displayed a red dot within two concentric circles, dubbed as the \textit{Target}, at a random location on screen. The recording procedure was divided into two phases, selection phase and fixation phase. In the selection phase, the participant had to look at the \textit{Target} and click on it. We monitored the distance between the \textit{Target} and mouse click position to verify whether participant selected it correctly. Once the user properly selected the \textit{Target}, the \textit{Target} image shrunk to a minimum size. This served as an affirmative feedback to the participant for their click action and also to obtain more precise ground truth point. Once the \textit{Target} reached to a minimum size, the fixation phase could be initiated by the participant by pressing \textit{space bar} on keyboard. During the \textit{Fixation phase}, participants were asked to fixate on the \textit{Target} while varying their head pose or facial appearance or position with respect to screen. Moreover, they were asked to be natural during fixation phase and the only constraint was to maintain their gaze at the \textit{Target} point. Our data recording happened in two phases spanning over a year. In the first phase, we allowed participants to start and stop the fixation at their own pace and in second phase, we terminated each fixation after a predefined time of 7 seconds. We turned the application background to black once the fixation starts and ensured that no other screen element including cursor got displayed during the fixation phase to minimize on-screen distraction. They were also given the option to terminate a fixation before the predefined duration by pressing 'Esc' key on keyboard in the event of any distraction or eye blinks. Participants got acquainted with the user interface and the recording software before the actual data recordings. Participants were located at different geographic locations and we instructed them to record in both indoor and outdoor locations. 

\begin{figure}[t]
\begin{center}
\includegraphics[width=\linewidth]{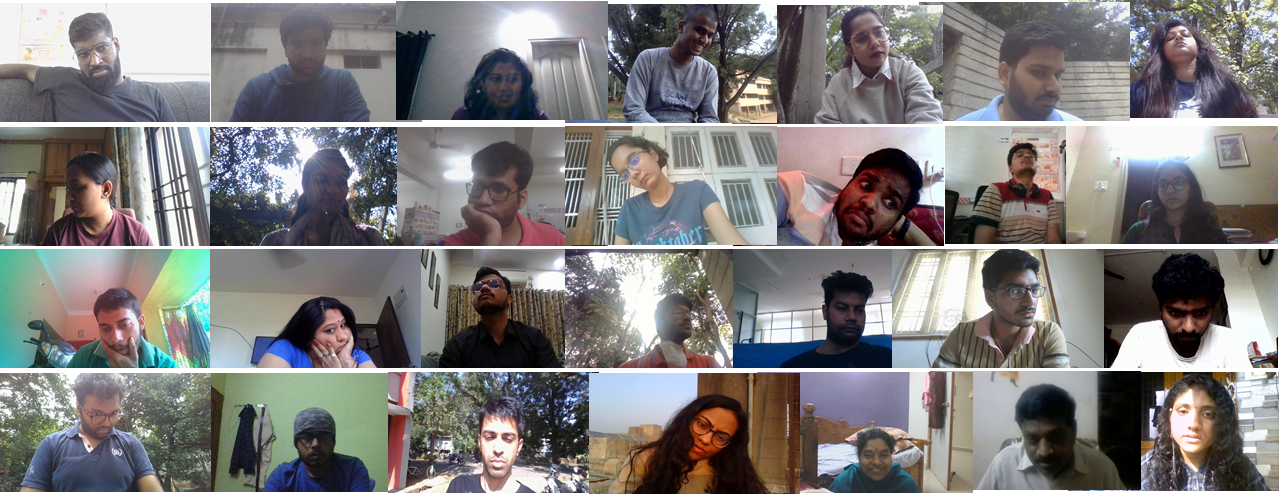}
\end{center}
   \caption{Representative Images from PARKS-Gaze Dataset. These samples demonstrate the variance across participants, location, head pose, appearance and illumination across the recordings}
\label{sampledatasetpic}
\end{figure}

\subsection{Sampling and Processing}

The number of sessions recorded by a single participant varied from 1 session to 23 sessions. Each session was predefined with 30 fixations, but many sessions had lesser fixations due to fixation terminations or session terminated by the participant. 
We recorded timestamps of participant's click on the \textit{Target}, space bar press and 'Esc' key press during the recording session. 
We observed from the recordings that all participants were static till the moment they began their fixation and they stopped orienting the head once the fixation was terminated. We believed that the predefined \textit{selection and fixation phase} design for data recording incorporated this head movement pattern and it allowed us to verify the attention of participant during the fixation phase. 
The mean duration over all fixation was 6.93 $\pm$ 1.26 sec. 
From our fixation recordings, we sampled at higher frequency (6 Hz) than the procedure reported for EYEDIAP in \cite{cheng2020gaze, zhang2017s, cheng2020coarse} (2 Hz) since the dynamics in terms of facial appearances and head pose in our dataset was higher. Further, EYEDIAP dataset contains both static and dynamic sessions whereas all fixations in the proposed dataset are dynamic by design. We obtained landmarks for sampled frames using OpenFace\cite{baltrusaitis2018openface}. We followed the procedure described in \cite{zhang18_etra} for image and gaze data normalization. The image normalization process cancels out the roll component of head orientation and places the image at a fixed distance from a virtual camera. Further, we transformed the PoG to a normalized space using extrinsic calibration \cite{takahashi2012new} of the laptop cameras, head pose and the head rotation matrix of the participant. 

\begin{figure}[t]
\begin{center}
\includegraphics[width=0.8\linewidth]{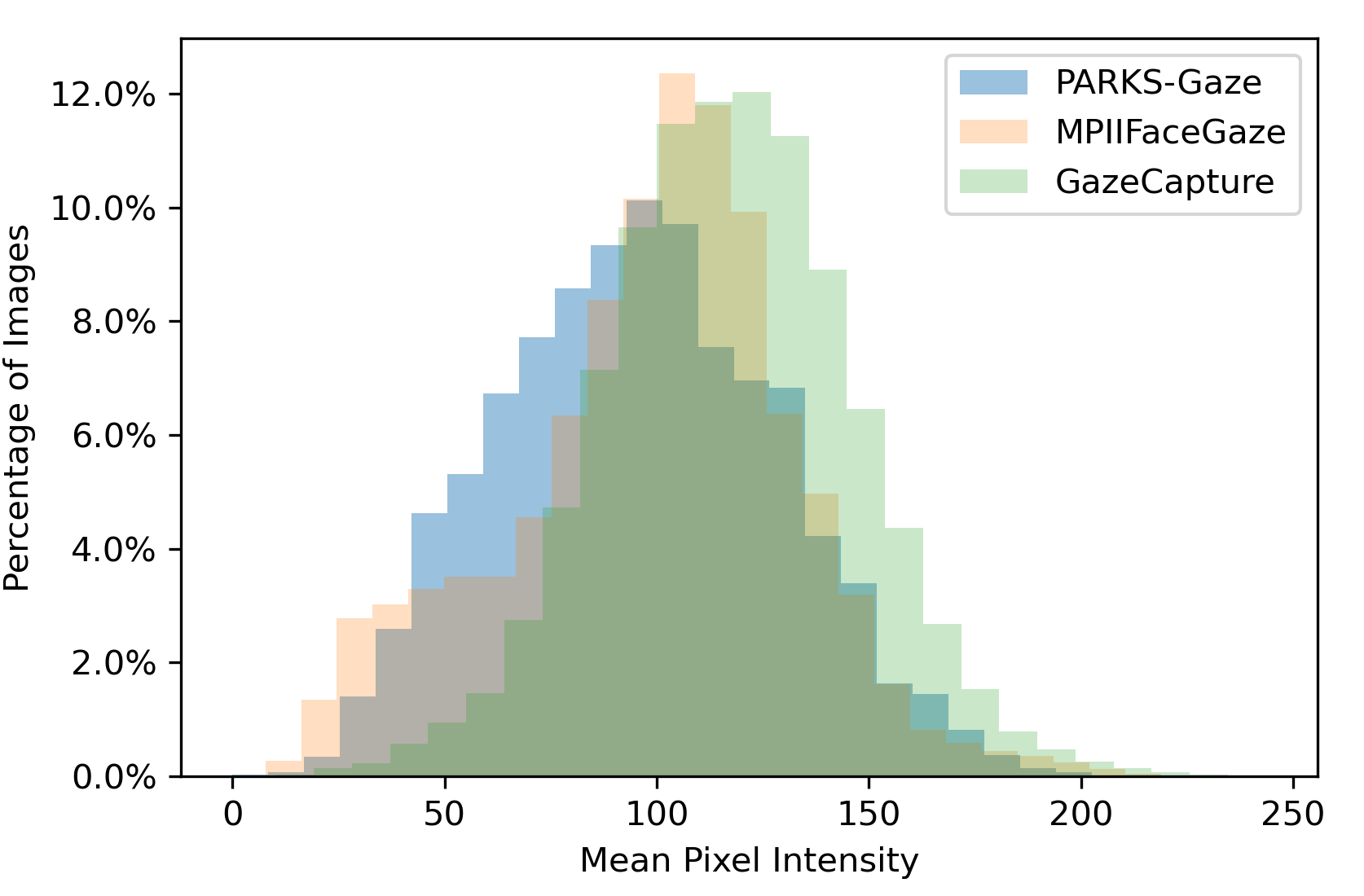}
\end{center}
   \caption{Mean Intensity Histogram of MPIIFaceGaze, GazeCapture and PARKS-Gaze datasets}
\label{illum_plots_together}
\end{figure}

\begin{figure*}[tbp]
  \centering
    \begin{subfigure}[b]{0.24\textwidth}
  \centering
    \includegraphics[width=0.95\linewidth]{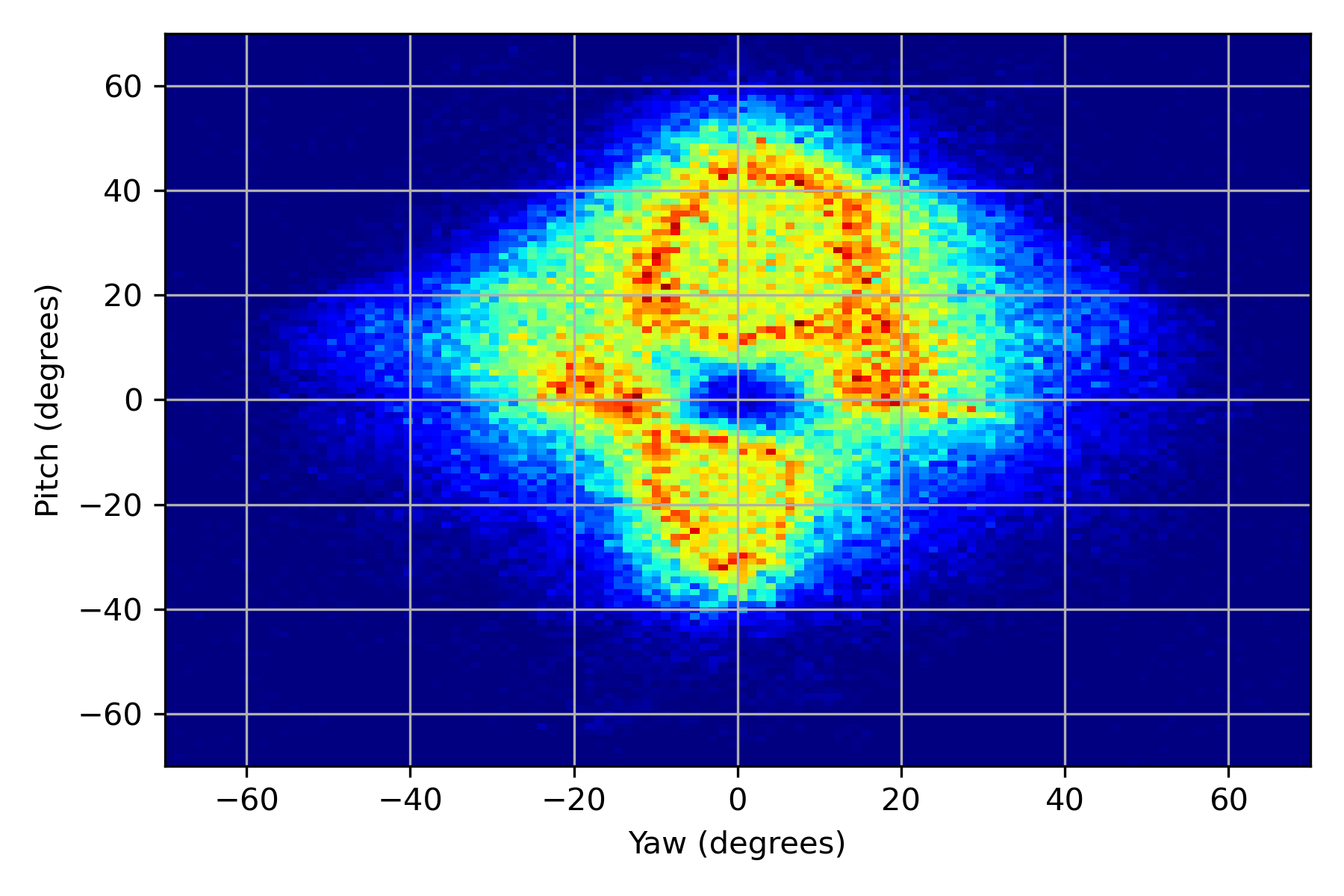}
       \caption{\label{PARKS-Gaze_HP}
    \textit{h} (PARKS-Gaze)}
    \end{subfigure}
  \begin{subfigure}[b]{0.24\textwidth}
  \centering
    \includegraphics[width=0.95\linewidth]{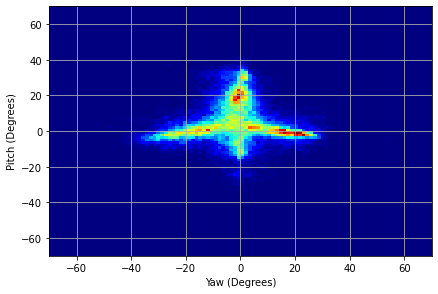}
    \caption{\label{mpiigaze_HP}
    \textit{h} (MPIIFaceGaze)}
    \end{subfigure}
  \begin{subfigure}[b]{0.24\textwidth}
  \centering
    \includegraphics[width=0.95\linewidth]{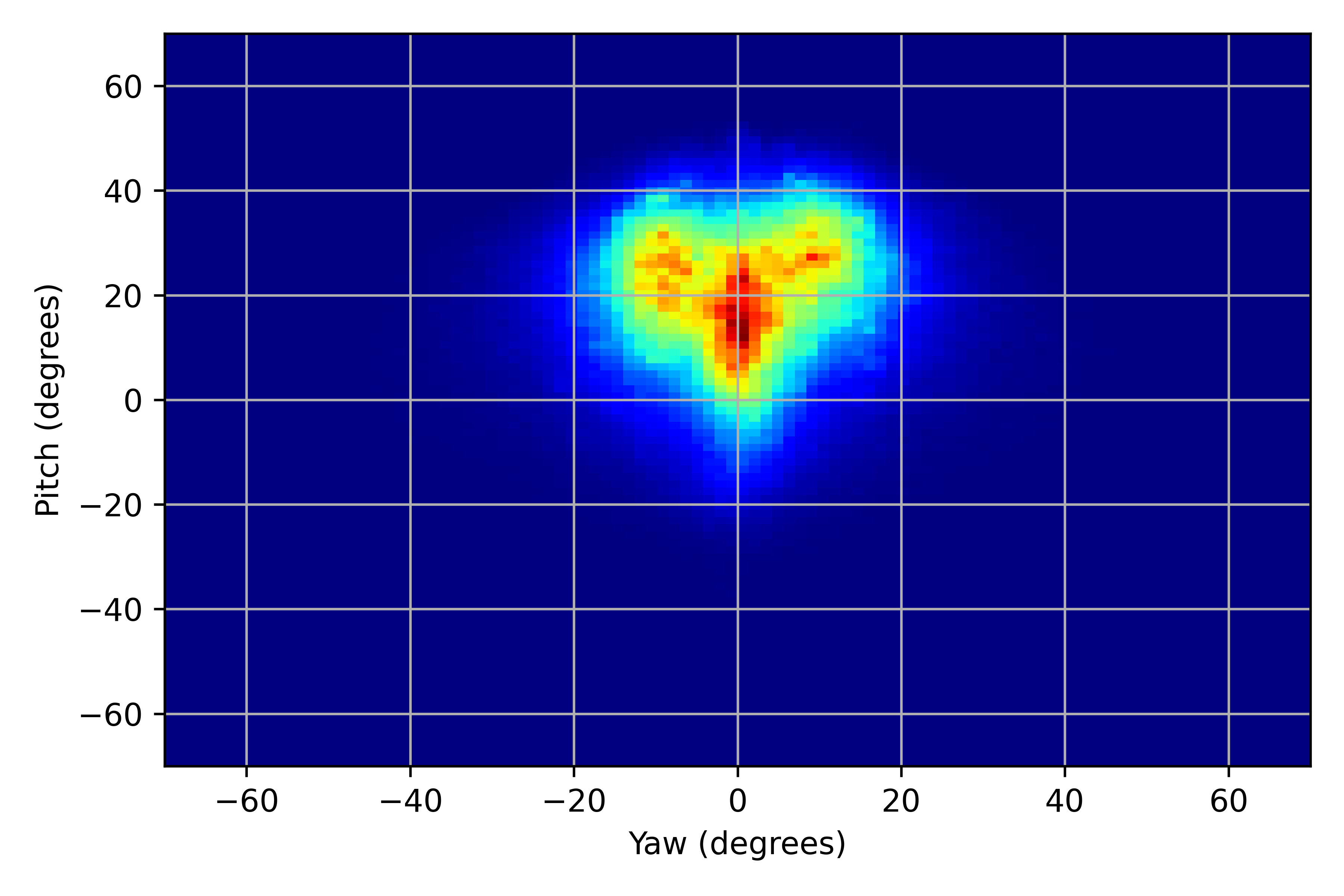}
    \caption{\label{gazecapture_hp}
   \textit{h} (GazeCapture)}
    \end{subfigure}
  \begin{subfigure}[b]{0.24\textwidth}
  \centering
    \includegraphics[width=0.95\linewidth]{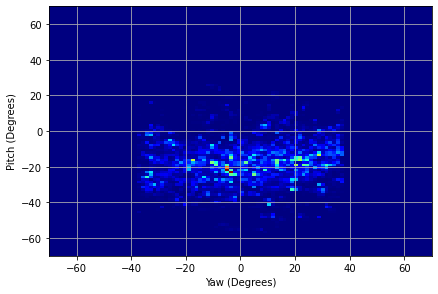}
    \caption{\label{rtgene_hp}
   \textit{h} (RT-GENE)}
    \end{subfigure}

  
    \begin{subfigure}[b]{0.24\textwidth}
  \centering
    \includegraphics[width=\linewidth]{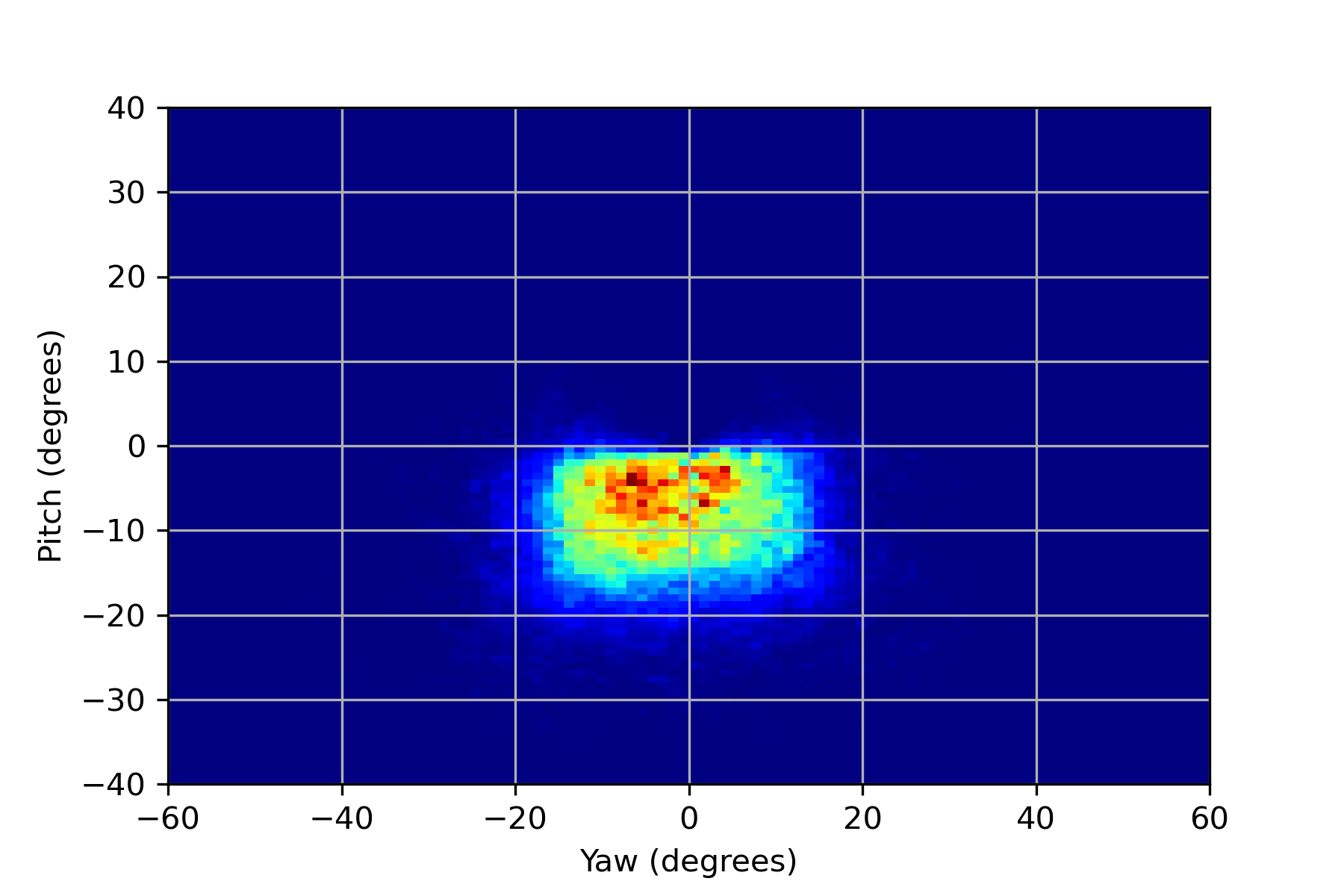}
       \caption{\label{parksgaze_gaze}
    \textit{g} (PARKS-Gaze)}
    \end{subfigure}
  \begin{subfigure}[b]{0.24\textwidth}
  \centering
    \includegraphics[width=0.9\linewidth]{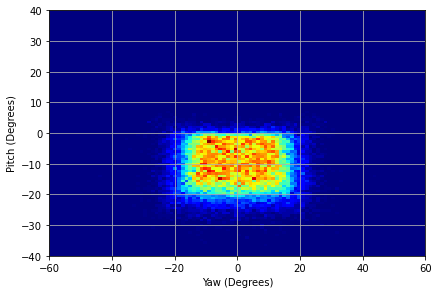}
    \caption{\label{mpiigaze_gaze}
    \textit{g} (MPIIFaceGaze)}
    \end{subfigure}
  \begin{subfigure}[b]{0.24\textwidth}
  \centering
    \includegraphics[width=0.9\linewidth]{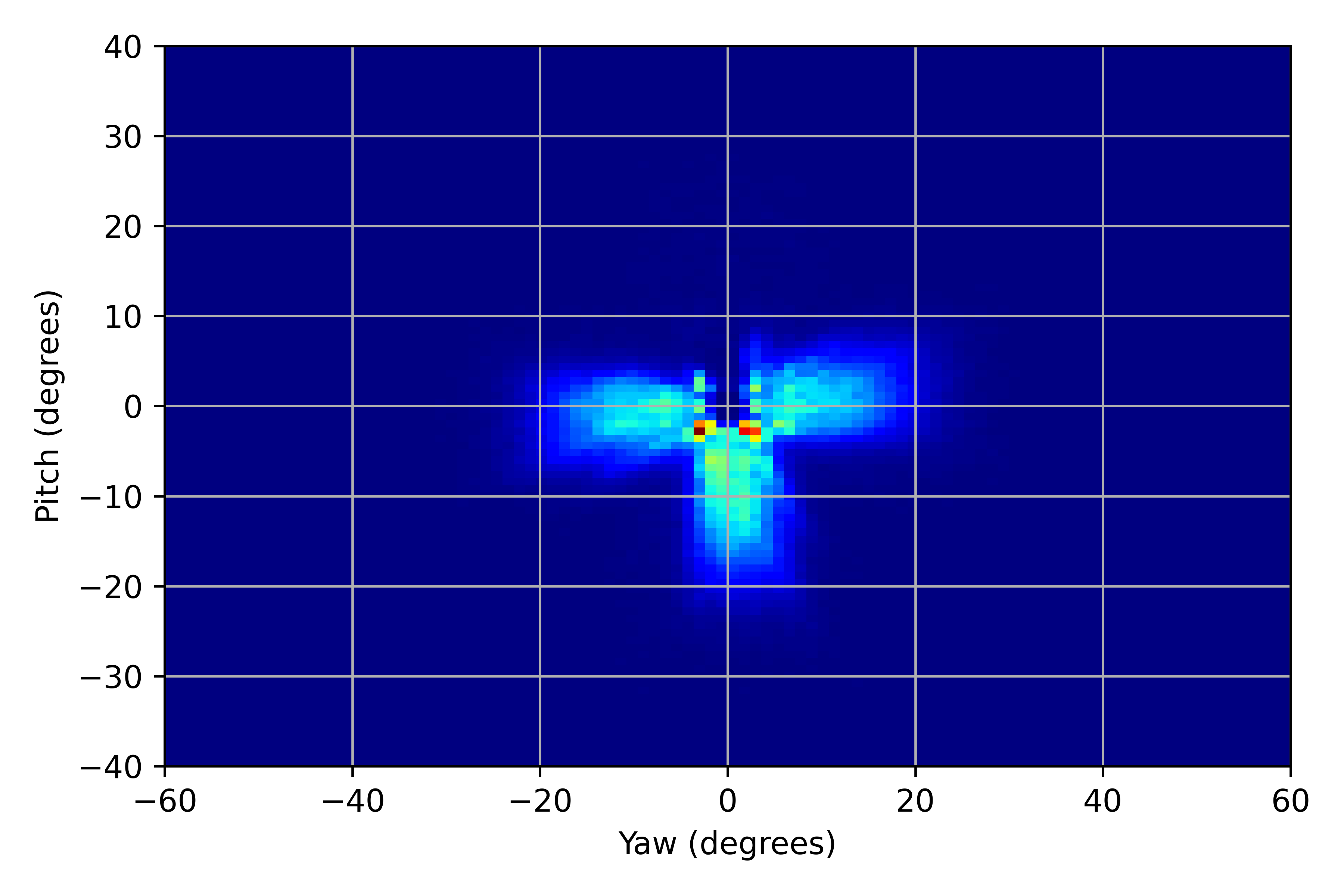}
    \caption{\label{gazecapture_gaze}
   \textit{g} (GazeCapture)}
    \end{subfigure}
  \begin{subfigure}[b]{0.24\textwidth}
  \centering
    \includegraphics[width=0.9\linewidth]{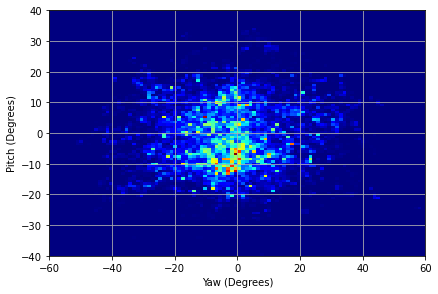}
    \caption{\label{rtgene_gaze}
   \textit{g} (RT-GENE)}
    \end{subfigure}
\caption{Distributions of head (\textit{h}) and gaze angles (\textit{g}) for PARKS-Gaze, MPIIFaceGaze, GazeCapture and RT-GENE Datasets}
\end{figure*}

\subsection{Dataset Characteristics}

Participants (17 Male and 11 Female; 16 with spectacles; Age: [19-56] years) belong to different states of India. We recorded 974 minutes of video recordings ( $\approx1.7$ Million images) over 328 sessions. 
The proposed dataset duration was larger than EYEDIAP (237 minutes) and data was collected in real-world conditions rather than in a lab setting. All sampled frames were manually verified and the frames with  eyes closed or user distraction from \textit{Target} were discarded. It was observed that the distraction can occur due to external noise while recording outdoors or the ambient environmental factors like wind or dust. Further, samples with confidence score less than 0.8 while detecting landmarks were discarded resulting in a final evaluation subset of 300,961 samples, larger than MPIIGaze \cite{zhang2017mpiigaze} and Gaze360 \cite{kellnhofer2019gaze360}. It may be noted that Gaze360 provides data at higher sampling rate of 8 Hz, yet contain less number of samples than the proposed dataset.  
A few representative images from the evaluation set were presented in Figure \ref{sampledatasetpic}. 

\subsubsection{Illumination}
We calculated mean pixel intensity of face crops of the proposed dataset and compared with MPIIFaceGaze \cite{zhang2017s} and GazeCapture \cite{krafka2016eye} datasets in Figure \ref{illum_plots_together}. It was observed that both MPIIFaceGaze (14.8\%) and the proposed PARKS-Gaze (15\%) contain similar percentage of images in low illumination range (Intensity $<$ 60) compared to GazeCapture (2.66\%). Further, it is evident from the plot that the proposed dataset has higher presence than the other two datasets on the left half of the intensity histogram where facial features are difficult to discern and gaze estimation approaches \cite{murthy2021appearance, cheng2020gaze} reported lower accuracy.

\subsubsection{Head Pose}
We illustrated the distribution of normalized head pitch and yaw for PARKS-Gaze, MPIIFaceGaze, GazeCapture and  RT-GENE datasets in Figures \ref{PARKS-Gaze_HP}, \ref{mpiigaze_HP}, \ref{gazecapture_hp}, \ref{rtgene_hp} respectively. It was observed that the range of head pose distribution of PARKS-Gaze dataset is $\pm60^{\circ}$ in both yaw and pitch directions, largest for an in-the-wild dataset. 
We noted that ETH-XGaze contain larger head pose range than the proposed dataset, but we observed that our participants followed \textit{head-pose-following-gaze} pattern and recorded data with naturally possible extreme head poses to gaze at the screen target. Moreover, head pose distribution of PARKS-Gaze is continuously distributed whereas ETH-X Gaze acquired discrete clusters of distribution, by the virtue of their multiple camera setup. Further, we achieved wider head pose range than other in-the-wild datasets like MPIIFaceGaze and GazeCapture. 

\begin{figure*}[tbp]
  \centering
    \begin{subfigure}[b]{0.32\textwidth}
  \centering
    \includegraphics[width=\linewidth]{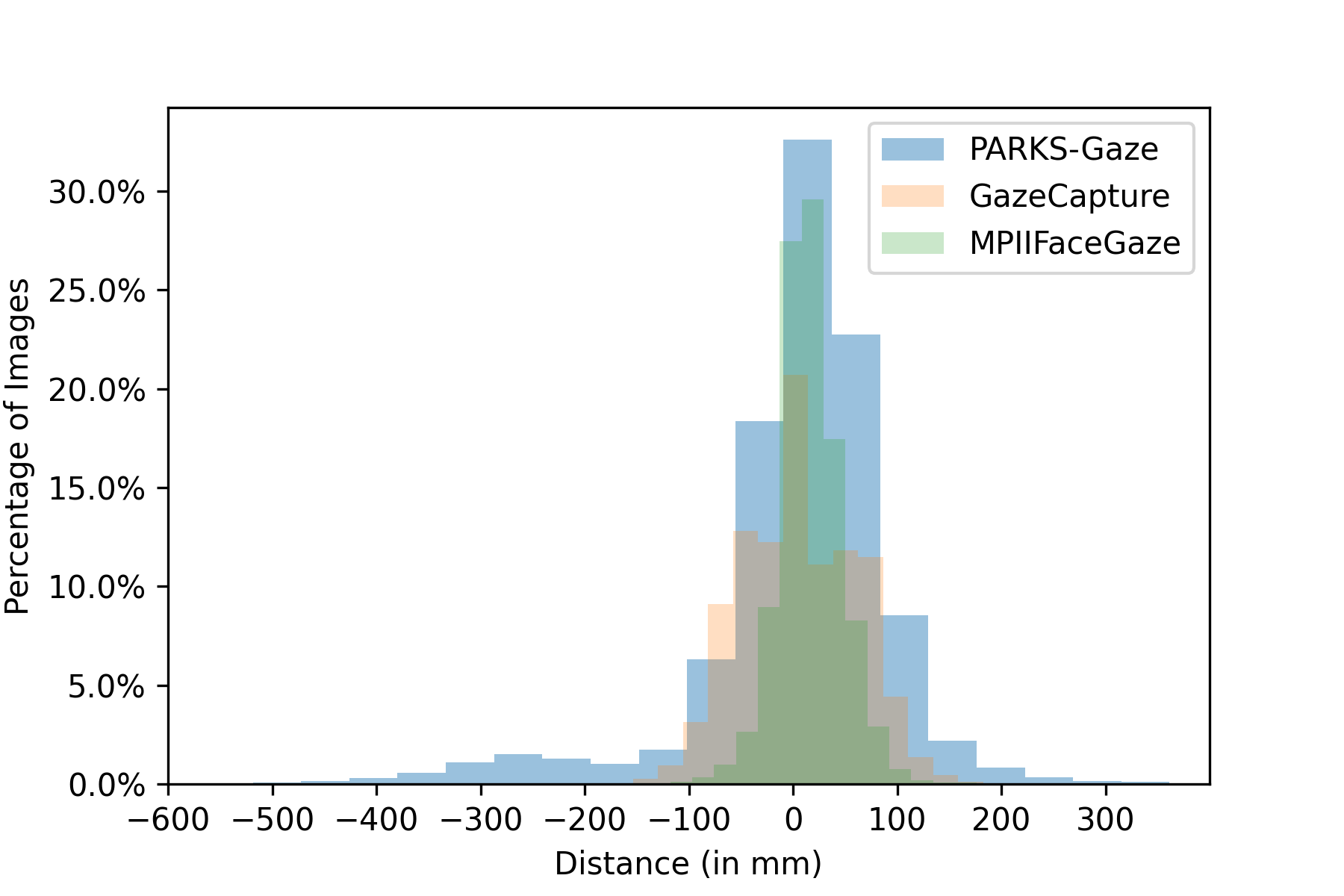}
       \caption{\label{distance_x}
    Location from Camera - \textit{X}}
    \end{subfigure}
  \begin{subfigure}[b]{0.32\textwidth}
  \centering
    \includegraphics[width=\linewidth]{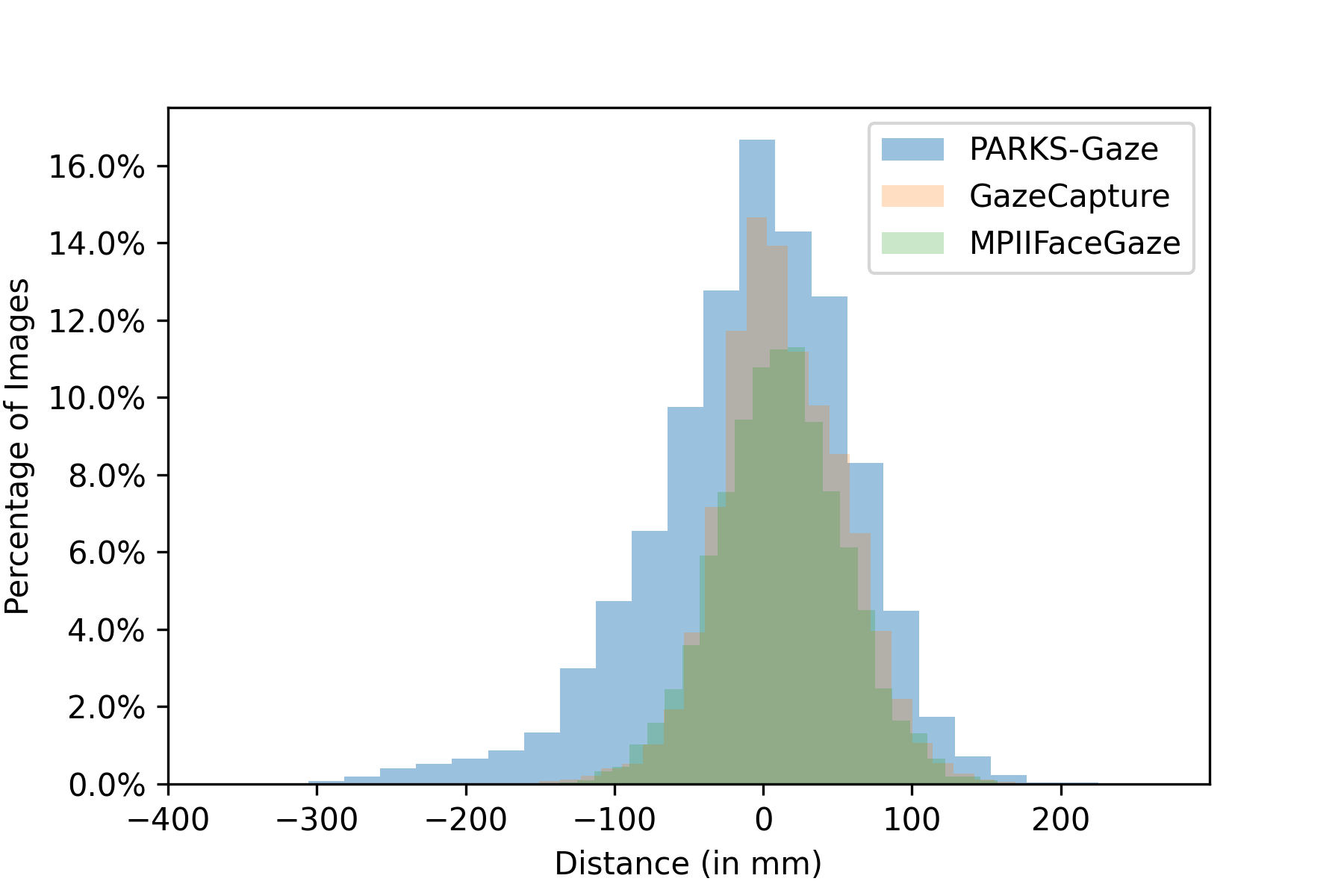}
    \caption{\label{distance_y}
    Location from Camera - \textit{Y}}
    \end{subfigure}
  \begin{subfigure}[b]{0.32\textwidth}
  \centering
    \includegraphics[width=\linewidth]{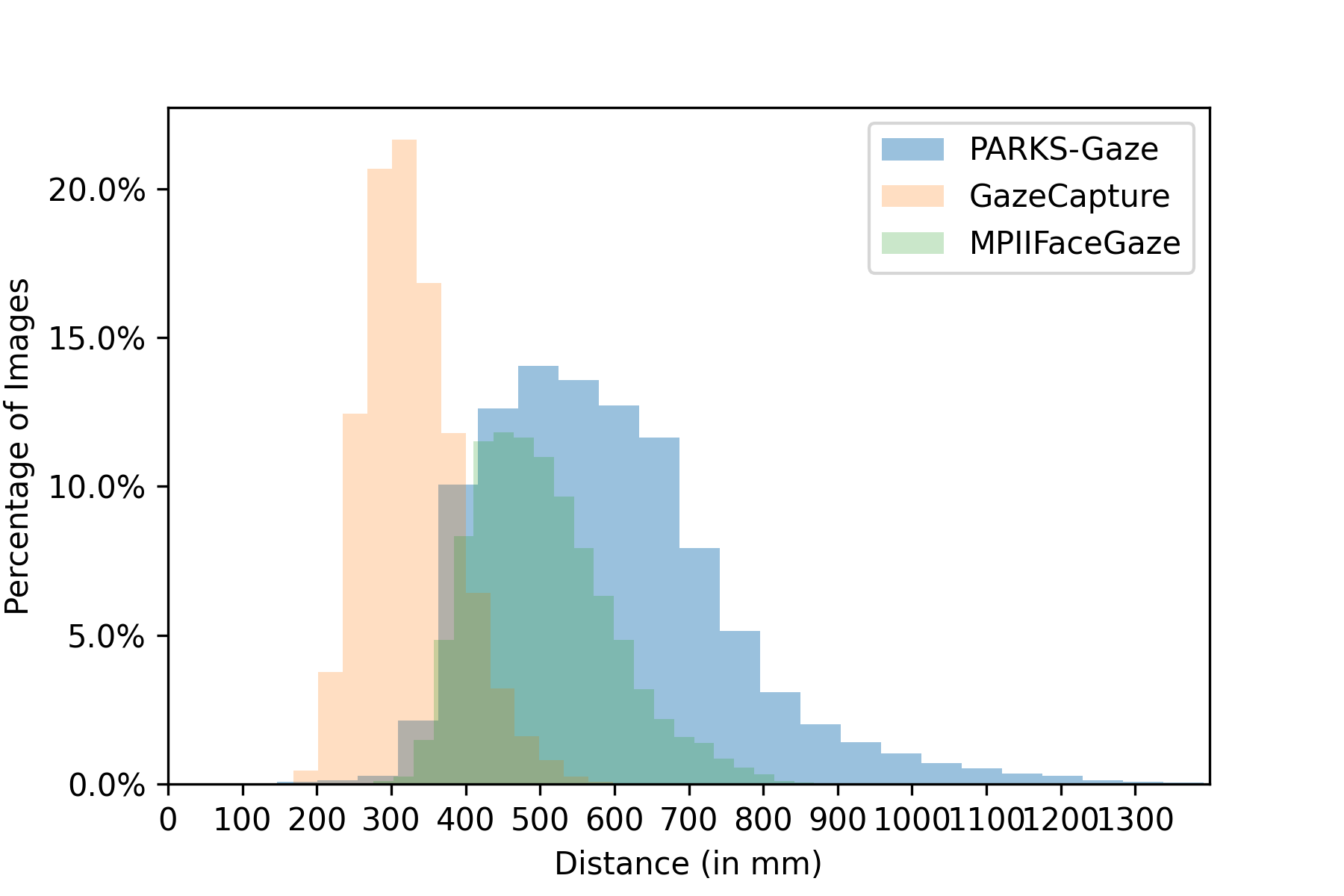}
    \caption{\label{distance_z}
   Location from Camera - \textit{Z}}
    \end{subfigure}
\caption{Histogram of participant's position in X, Y and Z directions in PARKS-Gaze, MPIIFaceGaze and GazeCapture Datasets}
\end{figure*}

In addition to face orientation, we compared the distribution of participant's position with respect to the camera across three datasets i.e., MPIIFaceGaze \cite{zhang2017s}, GazeCapture \cite{krafka2016eye} and the proposed PARKS-Gaze. We illustrated the same using histograms in Figures \ref{distance_x}, \ref{distance_y}, \ref{distance_z}. Across all three dimensions \textit{X}, \textit{Y} and \textit{Z}, we observed that the proposed dataset contain wider range of distances between the participant and the camera than the other two in-the-wild datasets in all three dimensions. Existing RT-GENE dataset \cite{fischer2018rt} which was collected in laboratory settings with RGB-D cameras focused on large distances between the participant and the camera in \textit{Z} direction with a range of 800-2800 mm, but provides little insight into the participant's position in Y-direction. We report widest range of participant positions in all three dimensions for an in-the-wild dataset recorded in interactive setting with a digital display. We observe that the range of GazeCapture in Z-direction is limited since they utilized handheld mobile displays and MPIIFaceGaze reported narrower range in X direction compared to other two datasets.

\subsubsection{Gaze direction}

We illustrated the normalized gaze angle distribution of PARKS-Gaze dataset ranging from $\pm30^{\circ}$, [10,-30$^{\circ}$] in yaw and pitch directions in Figure \ref{parksgaze_gaze} and we compared with MPIIFaceGaze, GazeCapture and RT-GENE datasets illustrated in Figures \ref{mpiigaze_gaze}, \ref{gazecapture_gaze}, \ref{rtgene_gaze} respectively. 
We observed that gaze region of RT-GENE dataset is wider in both yaw and pitch directions than other 3 datasets. We achieved similar gaze distribution as MPIIFaceGaze but the latter has higher presence at boundaries. 

\subsubsection{Precision}
The proposed dataset contain a total of 8432 fixations, each fixation with multiple images with variation in head pose, position of participant from the camera and facial appearance. This forms one of the major utility factors of our dataset. Interactive applications like gaze-typing \cite{majaranta2009fast}, gaze controlled wheel-chair interfaces \cite{singer2019see} require gaze estimation systems not only to be accurate but also to be precise for selection of screen elements. 
Since our dataset contain data from both indoor and outdoor conditions, 
it is useful for not only for precision evaluation of gaze estimation models, but also to train and build precise models that are robust under diverse environments.


\section{Experiments}\label{Experiments}


\begin{table*}
\centering
  \caption{Within-Dataset Evaluation on PARKS-Gaze and other datasets. $^{*}$ denotes the results from our implementations and we referred to the original works for other entries.}
  \label{withinval}
  \begin{tabular}{c|c|c|c|c} 
    \hline
    \toprule
    Models & PARKS-Gaze & MPIIFaceGaze & RT-GENE & EYEDIAP \\
    \midrule
    \midrule
    Spatial Weights CNN \cite{zhang2017s} & - & 4.8$^{\circ}$ & 10.0$^{\circ}$ & 6.0$^{\circ}$ \\
    Dilated-Net \cite{chen2018appearance} & 5.3$^{\circ}$$^{*}$ & 4.8$^{\circ}$ & - & - \\    FAR-Net \cite{cheng2020gaze} & - & 4.3$^{\circ}$ & 8.4$^{\circ}$ & 5.7$^{\circ}$ \\
    I2D-Net \cite{murthy2021i2dnet} & 5.6$^{\circ}$$^{*}$  & 4.3$^{\circ}$  & 8.4$^{\circ}$ & -\\
    CA-Net \cite{cheng2020coarse} & - & 4.1$^{\circ}$ & - & 5.3$^{\circ}$ \\
    AGE-Net \cite{murthy2021appearance} & \textbf{5.1$^{\circ}$}$^{*}$ & \textbf{4.1$^{\circ}$} & \textbf{7.4$^{\circ}$} & \textbf{5.2$^{\circ}$}$^{*}$\\

  \bottomrule
\end{tabular}
\end{table*}

\subsection{Within Dataset Evaluation}
We studied the performance of various existing gaze estimation models on the PARKS-Gaze dataset. We chose Dilated-Net \cite{chen2018appearance}, I2D-Net \cite{murthy2021i2dnet} and AGE-Net \cite{murthy2021appearance} since these models represent three levels of performance (4.8$^{\circ}$, 4.3$^{\circ}$ and 4.09$^{\circ}$ error respectively) on challenging MPIIFaceGaze dataset. We observed that other existing gaze estimation models \cite{cheng2020coarse, fischer2018rt, cheng2020gaze} reported similar performance as these methods. 
For our with-in dataset evaluation, we have separated our 28 participants into 3 groups, \textit{Train}, \textit{Validation} and \textit{Test} splits with 18, 4 and 6 participants in each group respectively. 
We ensured similar gaze and head pose distribution across three groups. 

As denoted in Table \ref{withinval}, AGE-Net achieved state-of-the-art performance on MPIIFaceGaze, EYEDIAP, RT-GENE and on the proposed PARKS-Gaze datasets. AGE-Net recorded a mean angle error of 5.17$^{\circ}$ across the 6 test participants of PARKS-Gaze. Since image resolution in both MPIIFaceGaze and PARKS-Gaze datasets is same, we posit that the higher with-in dataset validation error indicate diverse conditions and more number of challenging samples in the proposed dataset than MPIIFaceGaze. Since AGE-Net achieved superior performance across all datasets, we considered it for further evaluations.  

\subsection{Cross-Dataset Evaluation\label{Cross-datasetEval}}

We conducted pair-wise cross-dataset evaluations using MPIIFaceGaze, RT-GENE, EYEDIAP, GazeCapture, ETH-XGaze and the proposed PARKS-Gaze datasets. 
We followed the normalization procedure described in \cite{zhang18_etra} for all the above mentioned datasets. We used 960mm as the focal length of the virtual camera. The distance between the virtual camera and the image center for eye and face images is considered to be 600mm and 1000mm respectively. All the warped images are converted into grayscale and histogram equalized. 
Since authors of RT-Gene \cite{fischer2018rt} provided the normalized images, we resized and converted them to grayscale as per the model's requirements. In the case of EYEDIAP, we considered 14 participants with screen targets. We followed similar procedure followed in \cite{cheng2020gaze, zhang2017s} and sampled 1 frame for every 15 frames from VGA videos. We also considered the available frame annotations to eliminate frames with eyes closed or participant not looking at the \textit{Target} from our evaluation. Since GazeCapture do not have 3D gaze annotations, we utilized the normalized gaze annotations and head pose information provided by \cite{park2019few}. We considered 1176, 50 and 140 participants for training, validation and test splits.  For the ETH-XGaze dataset, we obtained landmarks using \cite{bulat2017far} and obtained normalized eye images from the given face images.

We summarized the results of our cross-dataset evaluation in Table \ref{crossval}. 
We ranked training datasets based on the mean angle error for each test dataset and assigned average ranking using the method described in \cite{zhang2020eth}. We observed training on PARKS-Gaze dataset leads to lowest test errors for RT-GENE and GazeCapture datasets and it obtained the best rank (1.6) followed by GazeCapture (1.8) among the 6 training datasets. Further, training on GazeCapture achieved lowest test error when tested on MPIIFaceGaze, ETH-XGaze and PARKS-Gaze datasets even though its gaze distribution is smaller than the test datasets. 

ETH-XGaze dataset reported 6.63$^{\circ}$, 7.97$^{\circ}$ and 11.65$^{\circ}$ test error on MPIIFaceGaze, GazeCapture and PARKS-Gaze dataset respectively. Since ETH-XGaze contained large number of participants and larger head angles, higher error on the proposed dataset than the MPIIFaceGaze and GazeCapture further corroborates our argument that PARKS-Gaze is more challenging than the other two datasets. Further, it reported large test errors on EYEDIAP and RT-GENE possibly due to the difference in image resolution between source and target datasets. 
We observed high errors as reported in \cite{zhang2020eth} when tested on ETH-XGaze due to the domain gap between other datasets and ETH-XGaze in terms of head pose and gaze angle distribution. It may be noted that the results demonstrated here in Table \ref{crossval} are superior to the ones obtained using the earlier version of the dataset \cite{lrd2022parks}.

\begin{table*}
\centering
  \caption{Cross-Dataset Validation Results (in Mean Angular Error)}
  \label{crossval}
  \begin{tabular}{c|c|c|c|c|c||c||c} 
    \hline
    \toprule
    \backslashbox{Train}{Test} & \cite{zhang2017s} & \cite{fischer2018rt} & \cite{funes2014eyediap} & \cite{krafka2016eye} & \cite{zhang2020eth} & PARKS-Gaze & Mean Rank \cite{zhang2020eth}\\
    \midrule
    \midrule
    MPIIFaceGaze \cite{zhang2017s}  & - & 18.9$^{\circ}$ & 17.5$^{\circ}$ & 7.2$^{\circ}$ & 40.3$^{\circ}$  &7.9$^{\circ}$ &  2.6\\
    RT-GENE \cite{fischer2018rt} & 18.2$^{\circ}$  & -  & \textbf{13.7$^{\circ}$} & 15.4$^{\circ}$ & 46.7$^{\circ}$ & 16.4$^{\circ}$ & 3.4  \\
    EYEDIAP \cite{funes2014eyediap} & 24.8$^{\circ}$ & 23.2$^{\circ}$ & -  & 22.5$^{\circ}$ & 61.0$^{\circ}$& 23.6$^{\circ}$ &  4.8 \\
    GazeCapture \cite{krafka2016eye} & \textbf{5.3$^{\circ}$} & 18.4$^{\circ}$ & 19.0$^{\circ}$ &  - & \textbf{32.3$^{\circ}$} & \textbf{7.2$^{\circ}$} &   1.8 \\ 
    ETH-XGaze \cite{zhang2020eth} & 6.6$^{\circ}$ & 35.1$^{\circ}$ & 20.2$^{\circ}$ & 7.9$^{\circ}$ & - & 11.6$^{\circ}$ & 3.8\\
    \midrule
    PARKS-Gaze & 5.9$^{\circ}$ & \textbf{16.9$^{\circ}$} & 16.5$^{\circ}$ & \textbf{6.5$^{\circ}$} &34.1$^{\circ}$    &- & \textbf{1.6}\\
  \bottomrule
\end{tabular}
\end{table*}

\subsection{Precision Experiments}

\begin{figure}[t]
\begin{center}
\includegraphics[width=\linewidth]{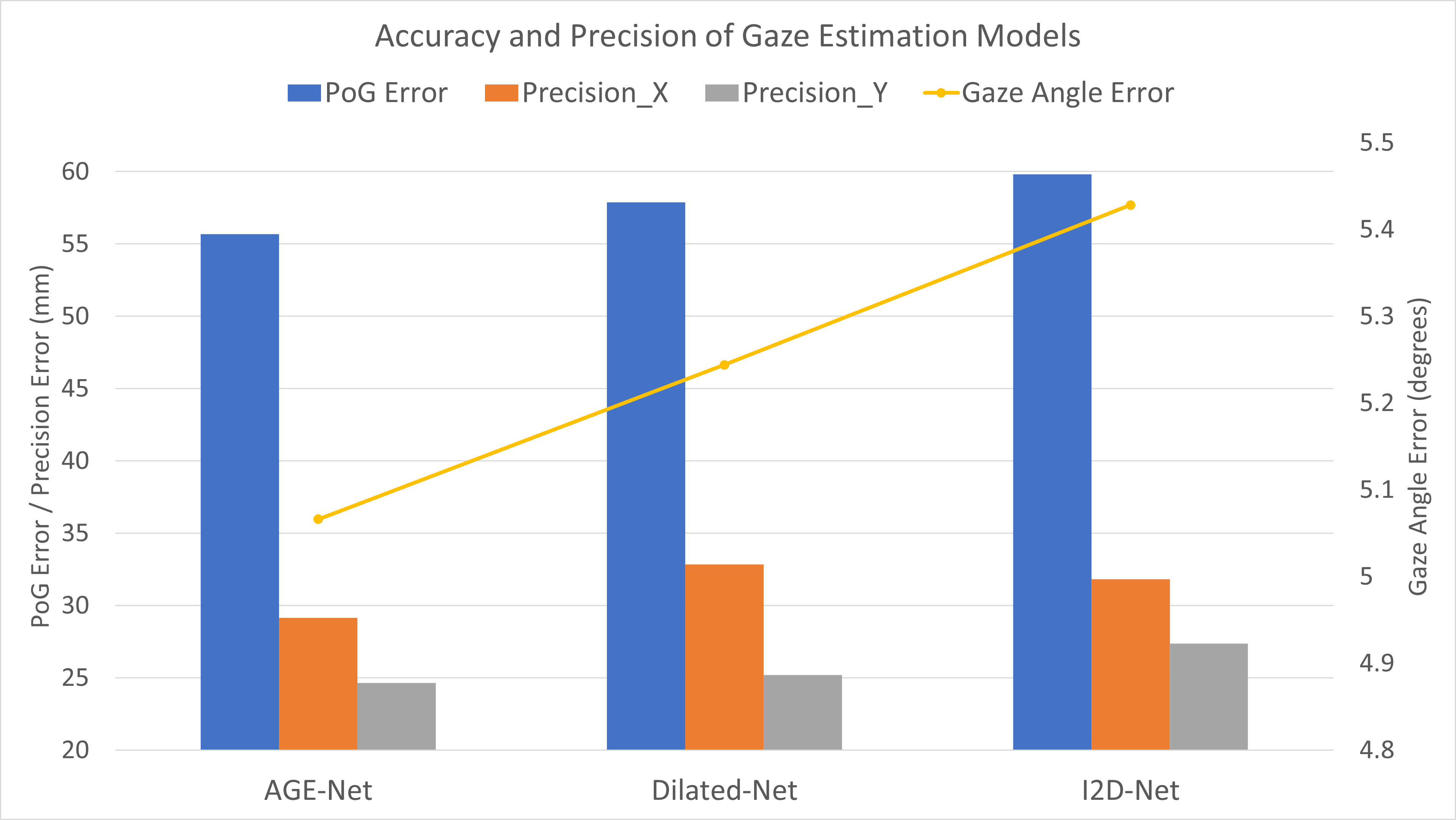}
\end{center}
   \caption{Accuracy and Precision of gaze estimation models}
\label{precisionaccuracy}
\end{figure}

Our literature review revealed that the evaluation of gaze estimation model's performance is predominantly confined to gaze angle errors. Such measurements do not inform us about the actual usability of such systems for interactive gaze-controlled applications. We undertook performance analysis of gaze estimation models at the PoG level investigating both accuracy and precision error. We performed these experiments in two fold. First, we performed precision analysis on the three models that we considered for our within-dataset validation. Second, we analyzed the effect of training dataset on precision error.

We defined the following experimental procedure to analyse precision error on an unseen participant. As described during within-dataset evaluation, we split our dataset into 3 groups with 18, 4 and 6 participants. 
We name these splits as \textit{Train}, \textit{Eval1}, \textit{Eval2}. We trained each model twice with \textit{Train} split using  \textit{Eval1} and \textit{Eval2} as the validation data and test data alternatively. This allowed us to investigate precision on a total of 10 unseen participants with 2504 fixations. 
We measured PoG error corresponding to each frame in a fixation using euclidean distance between the ground truth and the prediction in screen coordinate system measured in millimeters. We measured precision error using the standard deviation of all PoG predictions corresponding to each fixation. 
In Figure \ref{precisionaccuracy}, we presented the mean fixation error, an average value of PoG errors over all fixations of all 10 participants. Similarly, we presented Mean Precision\_X and Mean Precision\_Y which is an average value of all precision values computed over fixations for all participants in X and Y directions. We also illustrated the mean angle error of each model across all 10 test participants using a line plot. Two paired t-tests revealed that AGE-Net obtained statistically significant improvement over I2D-Net in precision in both X (t[9] = 3.02, p = 0.007) and Y (t[9] = 5.34, p = 2E-4) dimensions. Further, AGE-Net reported significant improvement over Dilated-Net in terms of precision in X-dimension (t[9] = 3.40, p = 0.003). 


\begin{figure}[t]
\begin{center}
\includegraphics[width=.81\linewidth]{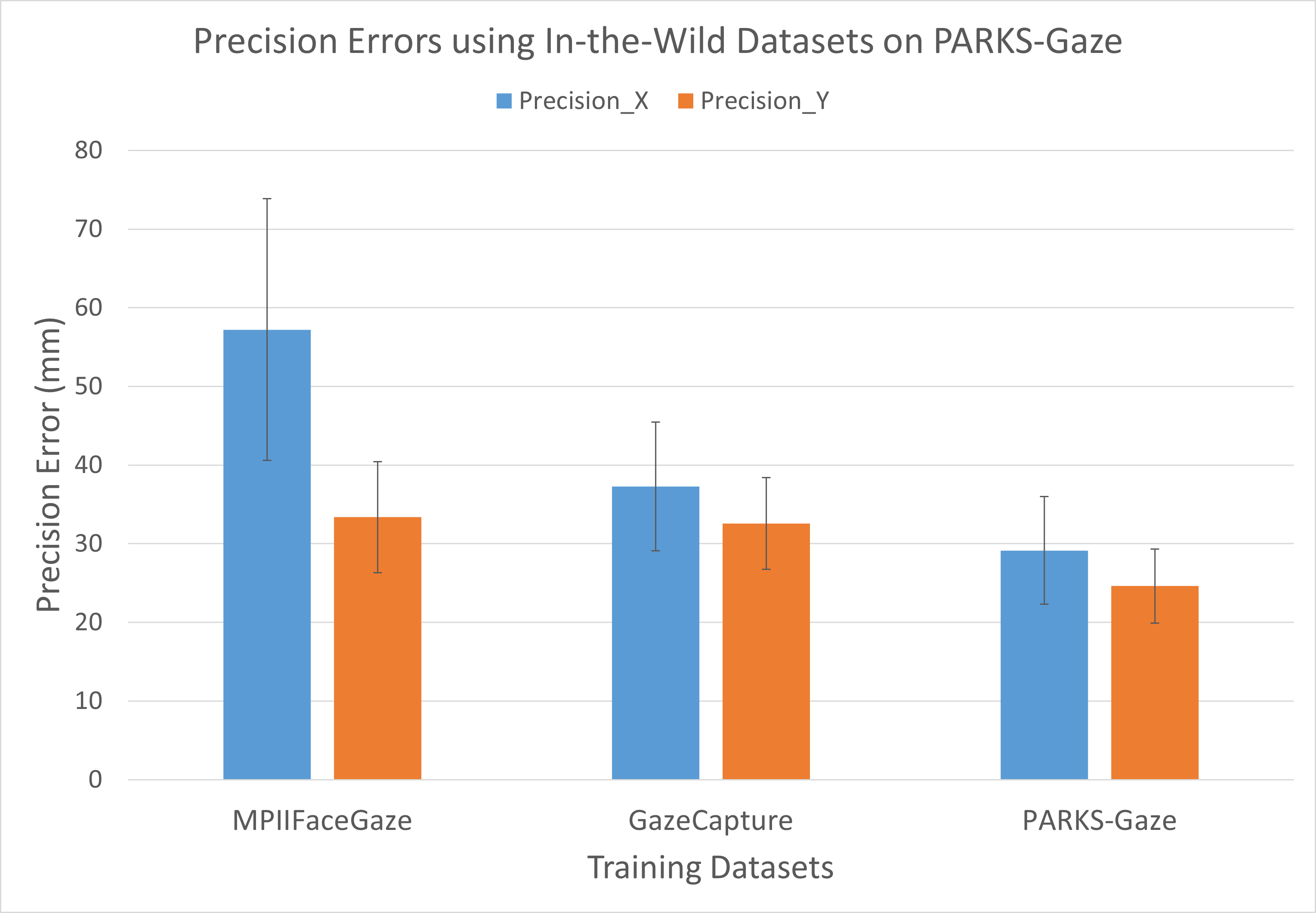}
\end{center}
  \caption{Effect of Training Dataset on Precision Error}
\label{precision_crossdataset}
\end{figure}

The cross-dataset evaluation reported in Section \ref{Cross-datasetEval} discusses only the accuracy of the model on an unseen participant from a different dataset. Here, we investigate the effect of training dataset on precision error under similar setting. For this purpose, we consider AGE-Net model as it reported superior precision over the other two models during within-dataset precision evaluation. 

We used \textit{Eval1} and \textit{Eval2} splits of the proposed PARKS-Gaze dataset and \textit{Test} split of GazeCapture dataset as test datasets for precision analysis. 
We discarded fixations from GazeCapture test split which have only one corresponding frame. The number of frames for remaining fixations ranged from 2 to 15. Since authors of GazeCapture did not provide extrinsic calibration parameters, \textit{{R, T}} of the camera with respect to the display, we attempted to generate them using the available information of screen points of the fixation and their corresponding representation in camera coordinate system. We were able to obtain extrinsic parameters for 23026 fixations. The \textit{{R, T}} are essential to map gaze angle predictions to screen coordinate system. 

On the \textit{Test} split of PARKS-Gaze dataset, training on MPIIFaceGaze resulted in a mean precision error of 57.2 $\pm$ 33 mm and 33.4 $\pm$ 14 mm while training on GazeCapture resulted in a mean precision error of 37.3 $\pm$ 16 mm and 32.6 $\pm$ 11 mm in X and Y directions respectively. Figure \ref{precision_crossdataset} illustrates the precision error obtained using MPIIFaceGaze, GazeCapture datasets. We further included results obtained when the model was trained on the \textit{Train} split of PARKS-Gaze dataset here for the sake of comparison. Two paired t-tests revealed that during this evaluation case, GazeCapture recorded statistically significantly lower precision error when compared to MPIIFaceGaze in X direction (\textit{X} - t[9] = 3.4, p = 4E-3) while no significant difference was observed for precision error in Y direction. 

Next, AGE-Net model trained on the \textit{Train} split of GazeCapture dataset reported a mean precision error of 8.2 $\pm$ 4.5 mm and 7.99 $\pm$ 3.5 mm in X and Y directions respectively when tested on its \textit{Test} split. On the \textit{Test} split of GazeCapture dataset, training on MPIIFaceGaze resulted in a mean precision error of 13.2 $\pm$ 6.5 mm and 10.4 $\pm$ 3.8 mm while training on the proposed PARKS-Gaze resulted in a mean precision error of 9.2 $\pm$ 4.2 mm and 8.07$\pm$2.8 mm in X and Y directions respectively. Two paired t-tests revealed that the proposed PARKS-Gaze dataset recorded statistically significantly lower precision error than MPIIFaceGaze on both X direction (\textit{X} - t[139] = 14.5, p = 1E-29) and Y direction (\textit{Y} - t[139] = 16.56, p = 6E-35). Hence, MPIIFaceGaze dataset is outperformed by both PARKS-Gaze and GazeCapture in two different scenarios in terms of precision errors. 

It is worth noting that we observed no significant difference between precision errors in Y-direction obtained using PARKS-Gaze dataset and GazeCapture dataset when tested on \textit{Test} split of GazeCapture dataset. Further, it may be noted that the combined mean precision error of PARKS-Gaze under cross-dataset setting (12.33 $\pm$ 4.9 mm) is only 0.8 mm lower than the precision-error under within-dataset setting using GazeCapture (11.52 $\pm$ 5.6 mm). GazeCapture training dataset has 1176 participants whereas PARKS-Gaze training split contained only 18 participants, yet the obtained precision errors under challenging cross-dataset setting are on-par to the errors obtained during a within-dataset evaluation. This indicates strong generalizability of the proposed dataset on unseen participants and its ability in generating precise gaze estimates. 

The mean precision error for fixations in \textit{Test} split of PARKS-Gaze with head pose variation $\leq5^{\circ}$ was observed to be 10mm and 9.8mm in X and Y directions. This represents the scenarios where a person might be dwelling on a button with an intention to select and these values decide the design choices while building gaze-controlled interfaces. A similar analysis on GazeCapture resulted in a  precision error of 13.5 mm, 14.8mm and on MPIIFaceGaze resulted in 15.7 mm, 13.7 mm precision error in X and Y directions. This indicate that the proposed dataset achieved a lower precision error on unseen participants even under limited head pose variation than existing in-the-wild datasets. Appearance-based gaze estimation is a function of many factors and the role of these factors in conjunction on the precision is not studied. 

\section{Effect of Head pose on Precision Error}

\begin{figure}[tbp]
  \centering
    \begin{subfigure}[b]{0.49\linewidth}
  \centering
    \includegraphics[width=\linewidth]{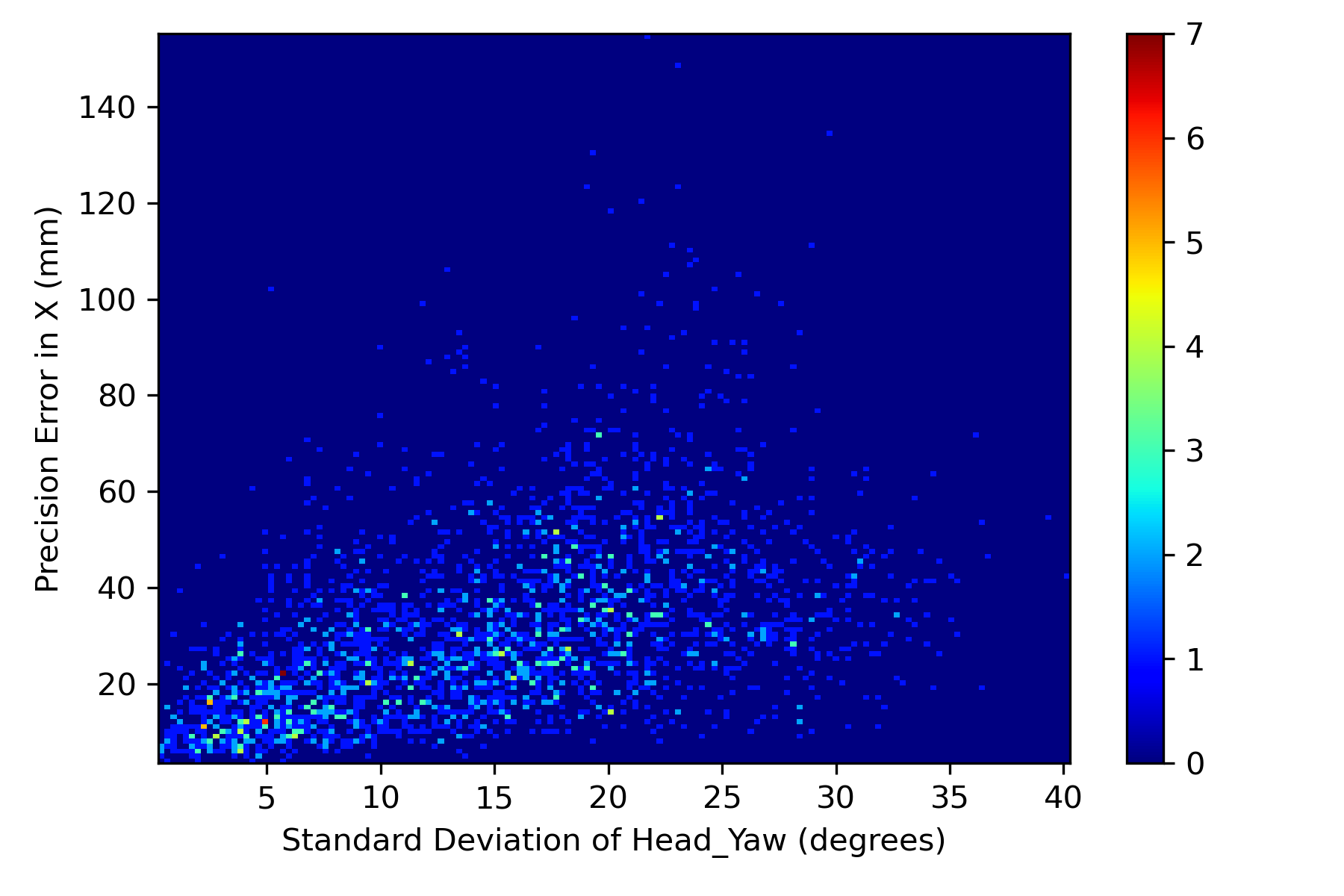}
       \caption{\label{HY_Vs_PX}
    Head Yaw Vs Precision\_X}
    \end{subfigure}
    \begin{subfigure}[b]{0.49\linewidth}
  \centering
    \includegraphics[width=\linewidth]{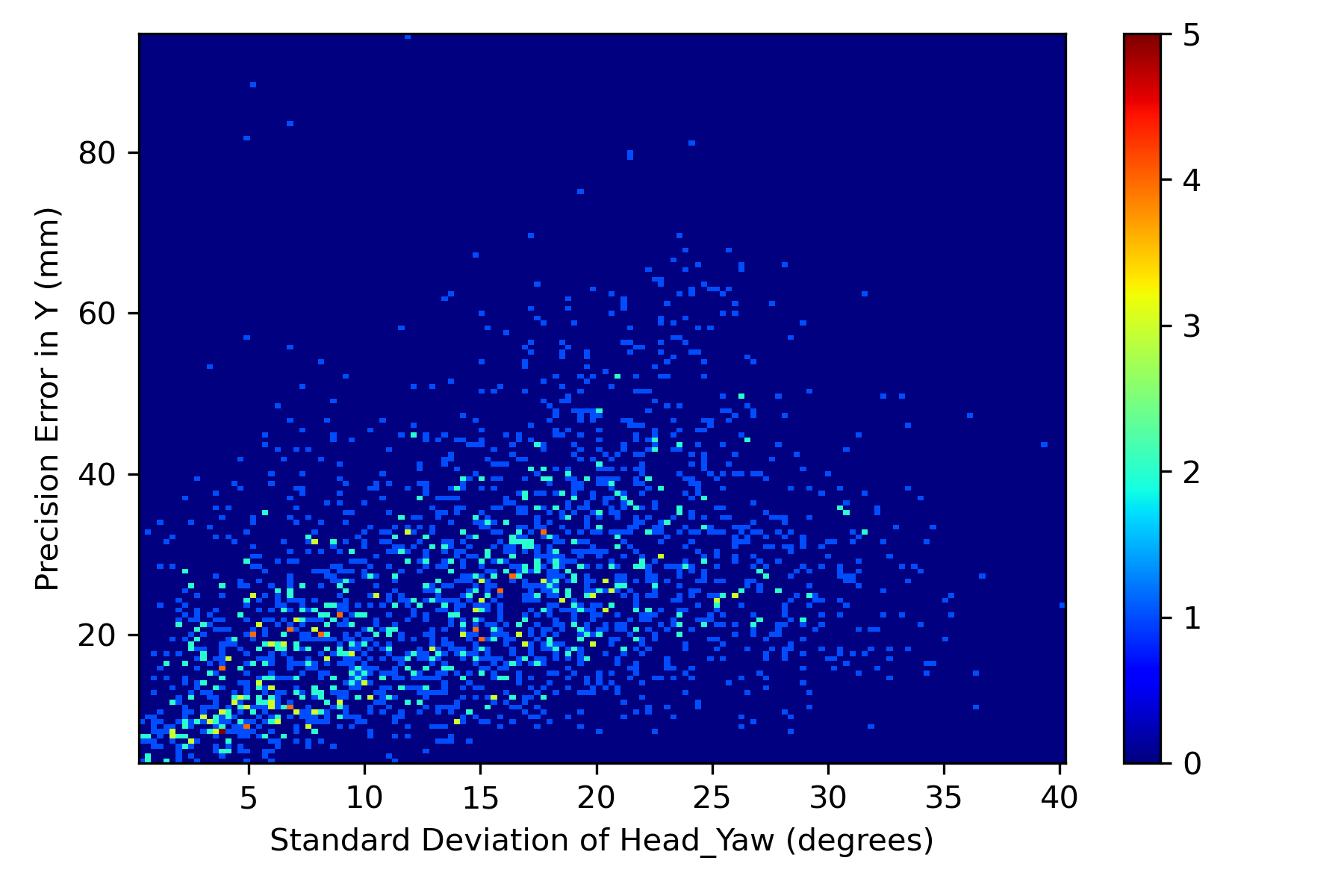}
       \caption{\label{HY_Vs_PY}
    Head Yaw Vs Precision\_Y}
    \end{subfigure}

  
    \begin{subfigure}[b]{0.49\linewidth}
  \centering
    \includegraphics[width=\linewidth]{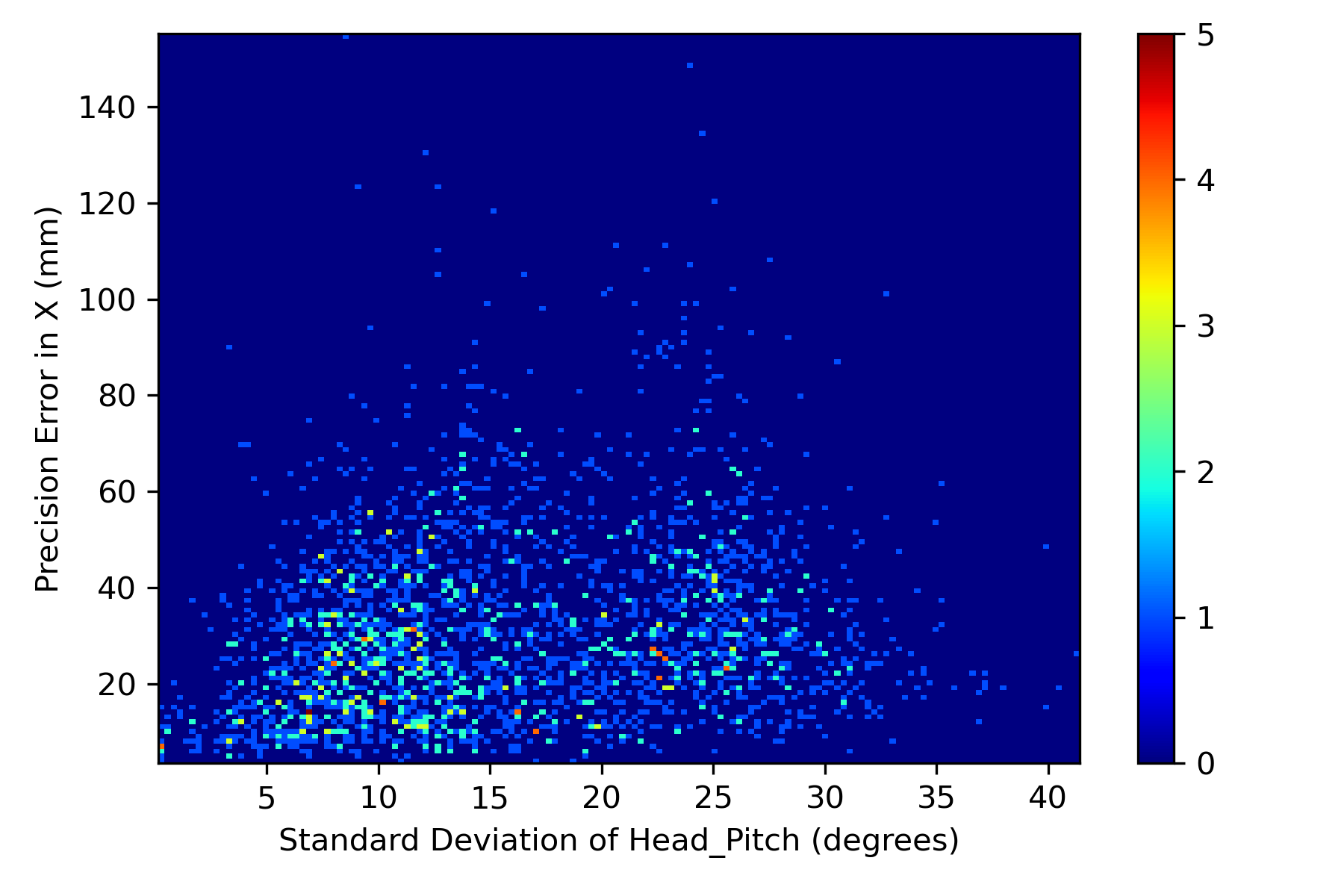}
       \caption{\label{HP_Vs_PX}
    Head Pitch Vs Precision\_X}
    \end{subfigure}
  \begin{subfigure}[b]{0.49\linewidth}
  \centering
    \includegraphics[width=\linewidth]{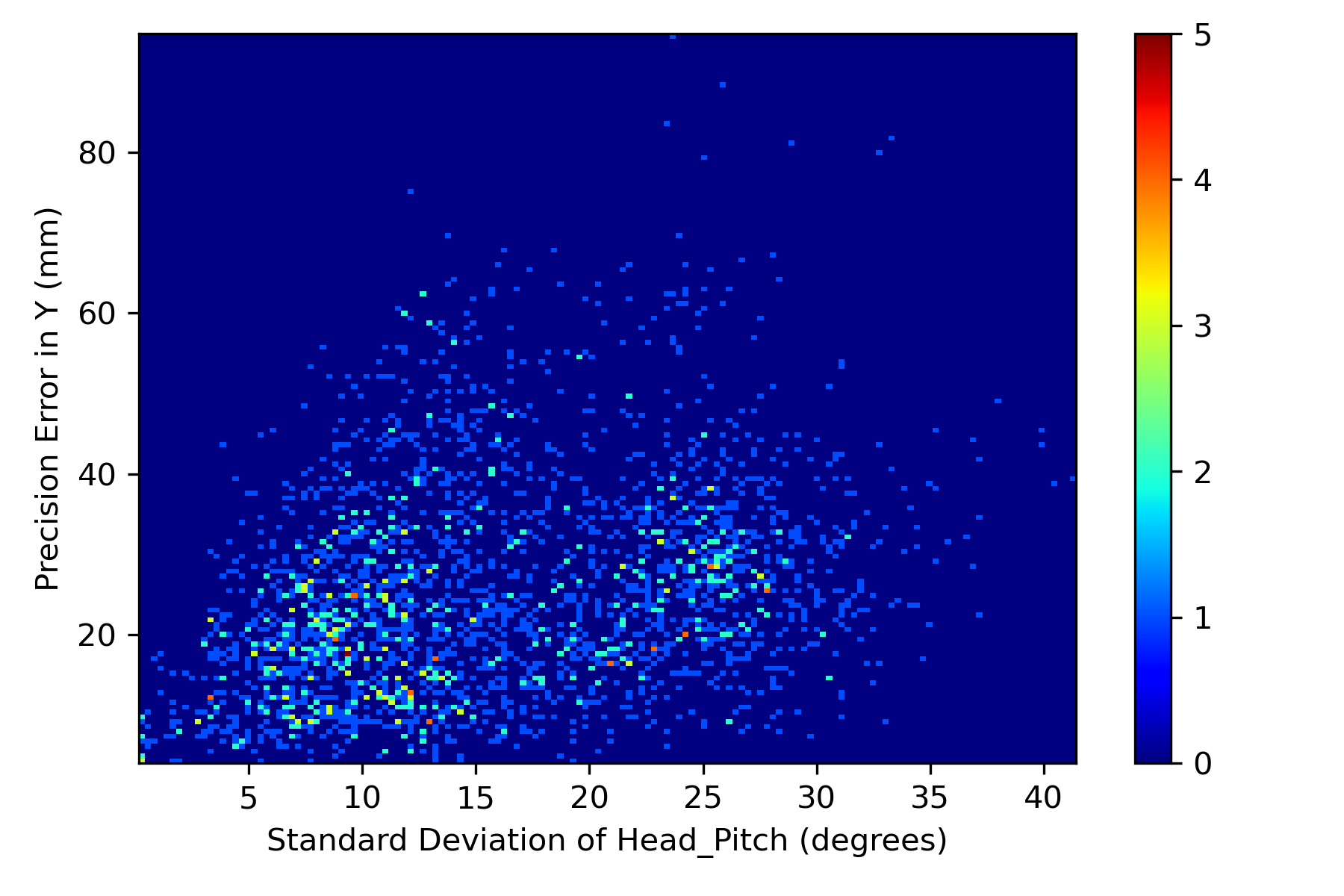}
    \caption{\label{HP_Vs_PY}
    Head Pitch Vs Precision\_Y}
    \end{subfigure}
\caption{Histograms of Precision Error in X and Y directions of each fixation w.r.t Standard deviation of Head Pitch and Yaw}
\end{figure}

In addition to evaluating precision error of models and the effect of training dataset on the precision error, 
we investigated the effect of head pose variation during each fixation on the precision error. We have considered 2504 fixations which have at least 5 normalized images as some images were discarded due to less landmarks confidence score. We computed the standard deviation of head yaw and pitch during each fixation and plotted histograms against the precision errors observed during the corresponding fixation in Figures \ref{HY_Vs_PX}, \ref{HY_Vs_PY}, \ref{HP_Vs_PX}, \ref{HP_Vs_PY}. Across all the fixations, we computed correlation between the standard deviation of head pose and the precision errors in both X and Y directions. It was observed that the mean correlation between head yaw with respect to precision in X (0.46) and Y (0.42) was moderately positive. For the case of standard deviation of head pitch, we found a weak correlation of 0.18 and 0.3 with precision in X and Y directions respectively. These moderate and weak Pearson's correlation coefficients is due to the fixations which have less head pose variation yet reported higher precision error, which can be observed in Figures \ref{HY_Vs_PX}, \ref{HY_Vs_PY}, \ref{HP_Vs_PX}, \ref{HP_Vs_PY}. We presented a qualitative analysis on such samples in the next section. 

\section{Qualitative Results - Precision Analysis}

\begin{figure*}[tbp]
  \centering
    \begin{subfigure}[b]{0.48\textwidth}
  \centering
    \includegraphics[width=0.9\linewidth]{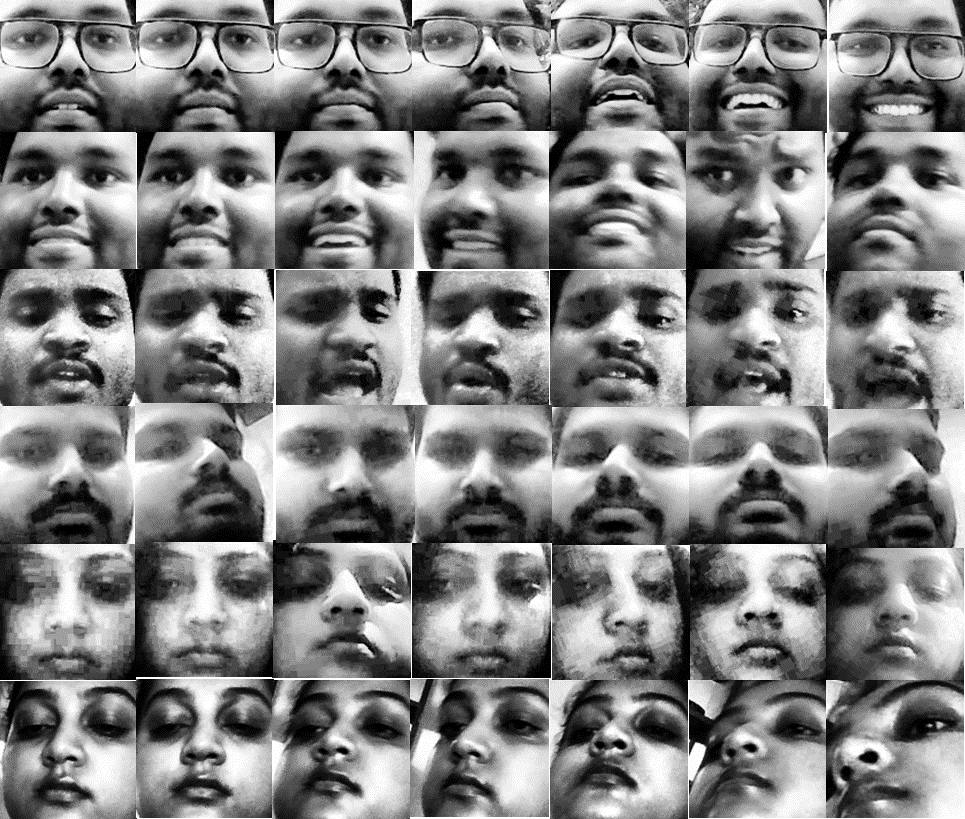}
       \caption{\label{Qual1}
     Fixations with head pose variation \textless 10$^{\circ}$}
    \end{subfigure}
  \begin{subfigure}[b]{0.48\textwidth}
  \centering
    \includegraphics[width=0.7\linewidth]{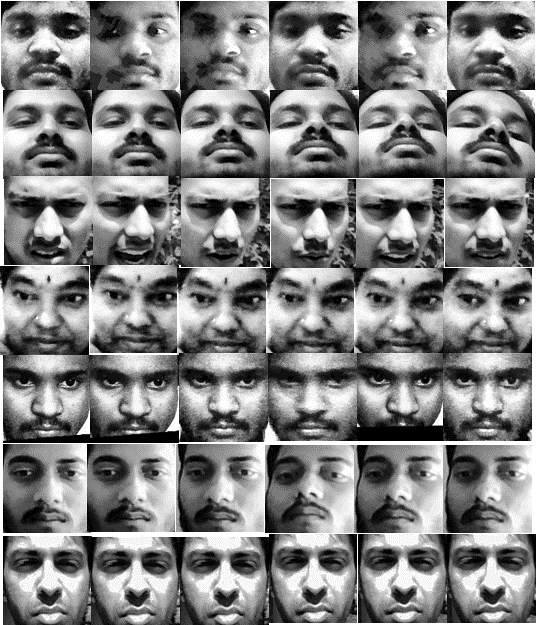}
    \caption{\label{Qual2}
     Fixations with head pose variation \textless 5$^{\circ}$}
    \end{subfigure}
\end{figure*}

We observed certain fixations which reported high precision error despite low variation in the head pose. We present the normalized face images from such fixations here for qualitative observations. 

For this purpose, we conducted our sampling in two-fold. First, we sampled the fixations with head pose variation in both pitch and yaw directions less than 10$^{\circ}$ and the precision errors in both X and Y directions higher than the mean value in the respective dimensions. Figure \ref{Qual1} presents the sequence of normalized face images from such instances. In this setting, we observed a maximum precision error in X and Y directions to be 49.5 mm and 48.2 mm respectively. These cases can be observed in the last two rows of Fig \ref{Qual1}. Visual inspection of these images allows us to consider the effect of facial expressions like smile (Row 1, 2), actions like talking (Row 3) on the gaze estimation performance thereby affecting the precision. Images from Row 5 has a distinct bright illumination effect and Row 6 reveals the variance during the cropping of face images at different head pose angles. 

On the other hand, we observed the images with precision errors in X and Y directions greater than 10 mm ($\approx$ 56px) with standard deviation in the head pose less than 5$^{\circ}$ in both pitch and yaw directions. Representative images from the fixations which fall under this category can be seen in Figure \ref{Qual2}. The maximum precision error in this case was observed to be 25mm and 34mm in X and Y directions respectively and the first two rows of Figure \ref{Qual2} represent these scenarios. Row 5 in Figure \ref{Qual2} presents the case of missing portion of face image which can possibly due to the closer position of the participant with respect to the camera. 
We inspected the camera frames corresponding to Row 6 and observed that participant moved back and forth from the camera. Camera frames from Row 1 revealed that even though pitch and yaw variation is limited, movement in roll direction is significant. It is interesting to note that roll component of head pose is not considered as the part of the gaze estimation as we cancel out during the normalization of images. Yet in this case, significant roll movement might be one of the factors for the poor precision. Images from Row 7 presents an interesting picture where the participant recorded least head movement yet the precision error is higher than 20 mm. This fixation was recorded in bright outdoor illumination conditions with 72K lux on the face region.


Summarizing, performance of appearance-based gaze estimation systems is affected by several factors like external illumination, position of the participant, facial expressions in addition to the head pose variation.

\textit{It is widely accepted that collecting real world datasets is challenging and the ones involving human participants is even trickier. The primal contribution of this dataset is to broaden the horizons formed by existing in-the-wild dataset in terms of head poses, illumination and intra-person appearance variations. Another key contribution of this work is generating the missing piece of knowledge on "precision" performance of gaze estimation systems. The proposed dataset was demonstrated to be challenging compared to other datasets while enabling the models to generalize well on other datasets. This indicate that we contribute an inclusive dataset along with unique scenarios which do not appear in other datasets. }

\section{Discussion}\label{Disc}

We presented PARKS-Gaze, a video dataset recorded in real world indoor and outdoor conditions focused on capturing wide range of head poses and intra-person appearance variations. Even though our dataset contained fewer number of participants than GazeCapture, we  illustrated the utility of our dataset in cross-dataset validation and precision experiments. Existing datasets like GazeCapture and MPIIFaceGaze reported ethnic diversity in their datasets yet our dataset containing all Indian participants reported best mean rank during the cross-dataset evaluation.  
Future works need to investigate the role of ethnic diversity in datasets on the generalization capability of gaze estimation models.  

During precision experiments, we showed that existing datasets reported higher precision errors when tested on proposed PARKS-Gaze dataset. Further, it was also demonstrated that the proposed dataset achieved on-par precision error with GazeCapture even under a cross-dataset setting. To evaluate GazeCapture and  PARKS-Gaze dataset for their cross-dataset precision on a common ground, no other dataset exist at present.  

We performed accuracy and precision measurements and investigated the effect of variation of head pose on precision. We shall investigate the effect of other factors like illumination and recording setting as part of our future work. We shall make this dataset available for research community with sequences of frames sampled at 6 Hz for each fixation. Future approaches can leverage techniques like Recurrent Neural Networks or Transformers \cite{vaswani2017attention} based Gaze estimation approaches for improved accuracy and precision.  


\section{Conclusion}\label{Conclusion}
We proposed PARKS-Gaze dataset, an appearance-based gaze estimation dataset which has broadened the boundaries of head pose and illumination characteristics of existing in-the-wild datasets. The final evaluation subset of our dataset consisted of 300,961 images and involve wide range of environments including outdoor settings. 
We demonstrated that proposed dataset encompasses more challenging samples and helps the model to generalize better in terms of both accuracy and precision than the existing in-the-wild gaze estimation datasets. 

{\small
\bibliographystyle{ieee_fullname}
\bibliography{egbib}
}

\end{document}